# Aggregation of Classifiers: A Justifiable Information Granularity Approach

Tien Thanh Nguyen[1], Xuan Cuong Pham[2], Alan Wee-Chung Liew[1], and Witold Pedrycz[3]

[1] School of Information and Communication Technology, Griffith University, Australia

[2] Department of Computer Science, Water Resources University, Hanoi, Vietnam

[3] Department of Electrical & Computer Engineering

University of Alberta, Edmonton, AB, T6R 2V4 Canada

**Abstract**: In this study, we introduce a new approach to combine multi-classifiers in an ensemble system. Instead of using numeric membership values encountered in fixed combining rules, we construct interval membership values associated with each class prediction at the level of meta-data of observation by using concepts of information granules. In the proposed method, uncertainty (diversity) of findings produced by the base classifiers is quantified by interval-based information granules. The discriminative decision model is generated by considering both the bounds and the length of the obtained intervals. We select ten and then fifteen learning algorithms to build a heterogeneous ensemble system and then conducted the experiment on a number of UCI datasets. The experimental results demonstrate that the proposed approach performs better than the benchmark algorithms including six fixed combining methods, one trainable combining method, Adaboost, Bagging, and Random Subspace.

1. **Introduction**

In supervised learning, the relationship between feature vectors and class labels of training observations is exploited to learn the discriminative decision model. As data gathered from different sources can vary quite substantially, a learning algorithm that achieve high accuracy on one dataset can perform less well on another dataset. Experiments have shown that there is no single learning algorithm that performs well on all data and it is difficult to know a priori which learning algorithm is suitable for a particular dataset. Hence, the research on how to combine several learning algorithms into a single framework to obtain a better discriminative decision model has generated a great deal of interest [1-3].

In many classification systems, the outputs usually reflect the probabilities of an observation belonging to given classes. However, in many practical situations, one may not be able to associate a precise probability with every event, particularly when only limited information is available. In this case, interval probabilities with lower and upper bounds provide a more general and flexible way to describe the uncertainty of the underlying knowledge [4]. Interval probability models have been successfully applied to many applications involving probabilistic and statistical reasoning, especially when there is a conflict between different sources of information [5].

In ensemble systems, each learning algorithm uses different methodology to learn base classifier on a given training set, thereby introducing uncertainty to the outputs. In ensemble learning, the meta-data of an observation reflects the agreements and disagreements between the different base classifiers. A combiner which can explicitly represent knowledge with uncertainty is therefore desirable. Several combiners that exploit this idea have been proposed, such as fuzzy integral in neural network [6] and Decision Template [7]. In this study, instead of dealing with precise numerical membership values like those encountered in traditional classification system, we propose a novel combining classifiers algorithm that captures the uncertainty in the outputs of base classifiers in an explicit manner using the notion of information granularity. Information granules and Granular Computing are directly attributed to the pioneering work by Zadeh [8-10] and further developed in [11-15]. Specifically, the prediction of base classifiers will be processed by justifiable information granularity to generate interval class memberships associated with class labels. As mentioned before, interval values are a flexible way to describe the uncertainty in the underlying knowledge. Therefore, the proposed algorithm will be more general than existing ensemble systems since it can output both interval values and crisp class memberships. Our experiments have confirmed that it performs significantly better than many existing ensemble systems.

The paper is organized as follows. In Section 2, we briefly discuss ensemble methods, with a focus on heterogonous ensemble systems. The concept of information justifiability in the design of information granules is also emphasized. In Section 3, a novel fixed combining method based on the idea of justifiable granularity is discussed. Experimental results are presented in Section 4 in which we compared the results of the proposed method to a number of benchmark algorithms on twenty one datasets. Finally, conclusions are presented in Section 5.

## 2. Prerequisites

### 2.1. Heterogeneous ensemble systems and fixed combining method

There are many taxonomies of ensemble method that focus on different factors and views at the ensemble systems [1, 16-18]. In [17], six strategies were introduced to build a sound combining system. The rationale behind these strategies is that "the more diverse the training set, the base classifiers, and the feature set, the better the performance of the ensemble system". The six strategies include (a) different initializations, (b) different parameter choices, (c) different architectures, (d) different classifiers, (e) different training sets, and (f) different feature sets. Two commonly used strategies encountered for ensemble systems are:

- Different training sets (also called Homogeneity scenario [19]): Generic classifiers are generated from applying the same learning algorithm onto different training datasets obtained from an original one. The outputs of these classifiers are then combined to produce the final decision. Several state-of-the-art ensemble methods in this category include AdaBoost [20], Bagging [21], Random Forest [22], and Random Subspace [23].

- Different classifiers (also called Heterogeneity scenario [19]): A set of different learning algorithms is used on the same training dataset to generate different base classifiers, a combiner then make decision from the outputs (called Level1 data or meta-data) of these classifiers [24-30]. This approach focuses more on the algorithms to combine meta-data to achieve higher accuracy than any single base classifier.

In this paper, we used several different classifiers learned from different learning algorithms on the same training set to construct a combining classifiers framework. There are two techniques to combine the outputs of different classifiers, namely fixed combining methods and trainable combining methods [18, 19]. Trainable combining methods work on the meta-data of training set to form the discriminative model. The important studies about trainable combining methods are based on the stacking algorithm, first proposed by Wolpert [31] and further developed in [24, 32]. In this algorithm, the original training set is divided into several disjoint parts of equal size. One part of the data plays the role of testing data in turn and the rest assume the role of training data during the training phase. The output of stacking is the posterior probability (called meta-data) that an observation belongs to a class according to each classifier. The common feature of stacking-based approaches is that the meta-data of the training set is trained again by a combiner to form the discriminative decision. Although exploiting the meta-data of training set to discover knowledge as done in trainable combining algorithms may enhance the classification accuracy, computational cost will also increase significantly. Several examples of trainable combining algorithms encountered in the literature are Variational Inference-based combiner [19], Multiple Response Linear Regression (MLR) [32], SCANN [33], and Decision Template [7].

In contrast, fixed combining methods do not take into consideration the label information in the meta-data of training set when combining. The advantage of applying fixed methods for ensemble system is that no training based on the class label of meta-data is needed; as a result, they are simple and less time-consuming than their counterparts. In fact, fixed combining methods are based on the Bayes decision model to integrate the predictions of classifiers associated with each class label. There are several popular fixed combining methods studied in the literature, namely Sum, Product, Majority Vote, Max, Min, and Median rules [34] (see Table 1). Of these, Vote and Sum are the most frequently used rules.

Let $\{y_m\}_{m=1,\dots,M}$ denotes the set of $M$ labels, $N$ denotes the number of observations, $K$ is the number of base classifiers. For an observation $\mathbf{x}$, $P_k(y_m|\mathbf{x})$ is the probability that $\mathbf{x}$ belongs to the class with label $y_m$ given by the $k^{th}$ classifier. There are two popular types of output for **x** for each $k = 1, \ldots, K$:

- *Crisp (Boolean) Label*: return only class label $P_k(y_m|\mathbf{x}) \in \{0,1\}$ and $\sum_m P_k(y_m|\mathbf{x}) = 1$
- *Soft Label*: return posterior probabilities that **x** belongs to classes, i.e. $P_k(y_m|\mathbf{x}) \in [0,1]$ and $\sum_m P_k(y_m|\mathbf{x}) = 1$

In this work, we focus only on the soft label. In this case, the posterior probability reflects the support of a class to an observation. The meta-data of an observation **x** is defined in the form of the following matrix:

$$\mathbf{L}(\mathbf{x}) = \begin{bmatrix} P_1(y_1|\mathbf{x}) & \cdots & P_1(y_M|\mathbf{x}) \\ \vdots & \ddots & \vdots \\ P_K(y_1|\mathbf{x}) & \cdots & P_K(y_M|\mathbf{x}) \end{bmatrix} \tag{1a}$$

While meta-data of all training observations, a $N \times MK$ posterior probability matrix, is defined as:

$$\mathbf{L} = \begin{bmatrix} P_1(y_1|\mathbf{x}_1) & \cdots & P_1(y_M|\mathbf{x}_1) & \cdots & P_K(y_1|\mathbf{x}_1) & \cdots & P_K(y_M|\mathbf{x}_1) \\ P_1(y_1|\mathbf{x}_2) & \cdots & P_1(y_M|\mathbf{x}_2) & \cdots & P_K(y_1|\mathbf{x}_2) & \cdots & P_K(y_M|\mathbf{x}_2) \\ \vdots & \ddots & \vdots & \ddots & \vdots & \ddots & \vdots \\ P_1(y_1|\mathbf{x}_N) & \cdots & P_1(y_M|\mathbf{x}_N) & \cdots & P_K(y_1|\mathbf{x}_N) & \cdots & P_K(y_M|\mathbf{x}_N) \end{bmatrix} \tag{1b}$$

TABLE.1. FIXED COMBINING RULES

| Rule name | Description |
|---|---|
| Sum rule | $\mathbf{x} \in y_t$ if $t = \arg\max_{m=1,\ldots,M} \sum_{k=1}^{K} P_k(y_m|\mathbf{x})$ |
| Product rule | $\mathbf{x} \in y_t$ if $t = \arg\max_{m=1,\ldots,M} \prod_{k=1}^{K} P_k(y_m|\mathbf{x})$ |
| Majority Vote rule | $\mathbf{x} \in y_t$ if $t = \arg\max_{m=1,\ldots,M} \sum_{k=1}^{K} \Delta_{km}$ <br> $\Delta_{kj} = \begin{cases} 1 & \text{if } j = \arg\max_{m=1,\ldots,M} P_k(y_m|\mathbf{x}) \\ 0 & \text{otherwise} \end{cases}$ |
| Max rule | $\mathbf{x} \in y_t$ if $t = \arg\max_{m=1,\ldots,M} \max_{k=1,\ldots,K} P_k(y_m|\mathbf{x})$ |
| Min rule | $\mathbf{x} \in y_t$ if $t = \arg\max_{m=1,\ldots,M} \min_{k=1,\ldots,K} P_k(y_m|\mathbf{x})$ |
| Median rule | $\mathbf{x} \in y_t$ if $t = \arg\max_{m=1,\ldots,M} \text{median}_{k=1,\ldots,K} P_k(y_m|\mathbf{x})$ |

## 2.2. Justifiable Information Granularity

If the probability distribution of data is known in advance, it is easy to represent the data by its distribution function. However, this information is usually unavailable in many real-world applications, and point estimates such as mean, median and skewness are often used to describe the data. Nevertheless, in many scenarios, pointwise information is less useful for subsequent reasoning [13]. Instead, information granularity explicitly models the inherent uncertainty present in the data. The concept of information granularity has been defined on many formal ways of describing information granules, such as sets ($A: \mathcal{X} \rightarrow \{0, 1\}$), fuzzy sets [35, 36] ($A: \mathcal{X} \rightarrow [0, 1]$), shadowed sets [37, 38] ($A: \mathcal{X} \rightarrow \{0, 1, [0, 1]\}$), and rough sets [39-41].

In this study, we aim to designing a single information granule to model the sample data **D** in the form of interval $\Omega = [a, b]$ in which $a$ and $b$ are lower and upper bounds of the interval, respectively. In that design, two intuitively compelling requirements need to be considered [13, 42-44]:

- Experimental evidence: The designed information granule $\Omega$ should reflect the existing experimental data so that the numeric evidence accumulated within the bounds of $\Omega$ attains the highest value. When the granule is formalized as a set (interval), the more data included within the bounds of the granule, the more legitimate this set becomes.
- Sound semantics: This requirement implies that the information granule should have well-defined semantics and exhibit high specificity. This implies that the smaller (more compact) the information granule (higher information granularity) is, the better (higher specificity) it is. For example, if the information granule comes in the form of an interval, the knowledge expressed as an interval [2, 4] is regarded to be more specific than the one residing within the interval [0, 10].

The principle of justifiable granularity is about constructing an information granule in the form of an interval to satisfy the two requirements outlined above. It is noted that two requirements mentioned above are only for the form of information granule proposed in this paper. In fact, there are several different approaches to formalize information granular such as in [45, 46] in which different requirements were used.

With regard to the first requirement, experimental evidence is quantified by counting the number of data points falling within the bounds of $\Omega$. If **D** consists of samples drawn from a certain probability density function $p$, the experimental evidence is given by the cumulative probability $\int_\Omega p(x)dx$. However, the distribution of **D** may not be known in advance or cannot be estimated reliably from a small number of observations in **D**. In this case, the experimental evidence can be determined by the cardinality of elements in **D** (denoted by C{**D**}) falling within the bounds of $\Omega$.

Meanwhile, the length of the interval is meaningful to model the specificity of the information granule $\Omega$ since shorter interval results in better specificity. To quantify this requirement, we use a continuous non-increasing function of the interval length. For instance, this function can be expressed in the form

$$f(u) = \exp(-\alpha u) \ (\alpha > 0) \tag{2}$$

in which $u = |a - b|$ is the length of interval $\Omega = [a, b]$, and $a$ and $b$ are the lower and upper bounds of the interval, respectively.

It is obvious that the two requirements are in conflict since increasing the cardinality will result in the reduction of the specificity. A compromise can be reached by using the product of these two functions:

$$\text{C}\{\mathbf{D}\} \times f(|a - b|) \tag{3}$$

To build the information granule $\Omega$ on a given dataset **D**, we select the median (denoted by $med(\mathbf{D})$) as the numerical representative of the experimental data. Then, $\Omega = [a, b]$ is formed by specifying its lower and upper bounds in which $a \leq med(\mathbf{D}) \leq b$. Since the upper and lower bounds are constructed independently, we only discuss the procedure to find $b$ ($a$ is determined in the same way). Based on (3) we have:

$$V(b) = \text{C}\{x_k \in \mathbf{D} \mid med(\mathbf{D}) \leq x_k \leq b\} \times f(|med(\mathbf{D}) - b|) \tag{4}$$

The optimal upper bound of the interval is determined by maximizing the values of $V(b)$ i.e.,

$$b_{opt} = \arg\max_{b \geq med(\mathbf{D}), b \in \mathbf{D}} V(b) \tag{5}$$

The optimal lower bound is found in the same manner

$$a_{opt} = \arg\max_{a \leq med(\mathbf{D}), a \in \mathbf{D}} V(a) \qquad (6)$$

where

$$V(a) = C\{x_k \in \mathbf{D} | a \leq x_k \leq med(\mathbf{D})\} \times f(|med(\mathbf{D}) - a|) \qquad (7)$$

The following algorithm summarizes the construction of information granule

---

**Algorithm 1: Constructing optimal lower and upper bound of information granule**

---

```
Input:         Sample data D = {x_k} and parameter α
Output:        Optimal lower and upper bound Ω = [a_opt, b_opt]

               Find med(D)

               (Finding the upper bound)

               For each b ∈ D, b ≥ med(D)

                   Compute V(b) by (4)

               End For

               b_opt = arg max_b V(b)

               (Finding the lower bound)

               For each a ∈ D, a ≤ med(D)

                   Compute V(a) by (7)

               End For

               a_opt = arg max_a V(a)

               Return [a_opt, b_opt]
```

---

## 3.  The Proposed framework

We now construct a combining method based on the concept of information granularity for the classification problem. In the proposed method, justifiable granularity will be applied to meta-data of observation to form the interval class memberships and then the predicted label is obtained via a translation to numerical class memberships. As the generated interval class memberships depends on $\alpha$, the performance of the method depends on $\alpha$ too. In the training process described in the Algorithm 2, we first introduce a method to find the optimal value of $\alpha$ from a set $\mathcal{A}$ by exploiting the meta-data of training observations. In this algorithm, we divide the training set $\mathcal{D}$ into $T$ disjoint parts $\{\mathcal{D}_1, \dots, \mathcal{D}_T\}$, where $\mathcal{D} = \mathcal{D}_1 \cup \dots \cup \mathcal{D}_T$ and $|\mathcal{D}_1| \approx \dots \approx |\mathcal{D}_T|$, and their corresponding $\{\mathcal{D}^{-1}, \dots, \mathcal{D}^{-T}\}$ in which $\mathcal{D}^{-t} = \mathcal{D} - \mathcal{D}_t$. Then, T-fold CV is applied onto training set $\mathcal{D}$ such that the meta-data of observations in $\mathcal{D}_i$ is obtained by classifiers generated by learning the $K$ learning algorithms on the associated part $\mathcal{D}^{-i}$ (denoted by $BC_k^{-i}$ in Algorithm 2). The meta-data of all training observations in $\mathcal{D}$ form a $N \times MK$ matrix $\mathbf{L}$ as in (1b) in which the $n^{th}$ row of $\mathbf{L}$ is the prediction (meta-data) for training observation $\mathbf{x}_n$. For each $\mathbf{x}_n$, we apply the principle of justifiable granularity to its meta-data to construct the interval membership values and then predict the class label of $\mathbf{x}_n$ based on a discriminative decision model operating on the intervals. In (1a), the $m^{th}$ column is the output of classifiers for predicting $\mathbf{x}_n$ to be in the $m^{th}$ class. For each value of $\alpha$ in $\mathcal{A}$, we apply Algorithm 1 on meta-data $\mathbf{x}_n$ to obtain the interval class memberships $\left[\underline{P}(y_m|\mathbf{x}_n), \overline{P}(y_m|\mathbf{x}_n)\right]$, $(m = 1, \dots, M)$ (8)

$$\mathbf{L}(\mathbf{x}_n) \coloneqq \begin{bmatrix} P_1(y_1|\mathbf{x}_n) & \cdots & P_1(y_i|\mathbf{x}_n) & \cdots & P_1(y_M|\mathbf{x}_n) \\ \vdots & \ddots & \vdots & \ddots & \vdots \\ P_K(y_1|\mathbf{x}_n) & \cdots & P_K(y_i|\mathbf{x}_n) & \cdots & P_1(y_M|\mathbf{x}_n) \end{bmatrix}$$

$$\underbrace{\left[\underline{P}(y_1|\mathbf{x}_n), \overline{P}(y_1|\mathbf{x}_n)\right]}_{interval\ for\ class\ 1} \cdots \underbrace{\left[\underline{P}(y_i|\mathbf{x}_n), \overline{P}(y_i|\mathbf{x}_n)\right]}_{interval\ for\ class\ i} \cdots \underbrace{\left[\underline{P}(y_M|\mathbf{x}_n), \overline{P}(y_M|\mathbf{x}_n)\right]}_{interval\ for\ class\ M} \quad (8)$$

Reasoning can be done on the interval membership values, e.g. using interval arithmetic [47], to form the final classification result. In this paper, we introduce a transformation from intervals in (8) to numerical class memberships using the following expression:

$$\text{NCM}(\mathbf{x}_n \in y_m) = g\left(\underline{P}(y_m|\mathbf{x}_n), \overline{P}(y_m|\mathbf{x}_n)\right) \times h\left(\left|\overline{P}(y_m|\mathbf{x}_n) - \underline{P}(y_m|\mathbf{x}_n)\right|\right) \quad (9)$$

where $\text{NCM}(\mathbf{x}_n \in y_m)$ denotes numerical class memberships that $\mathbf{x}_n$ belongs to class $y_m$, $g\left(\underline{P}(y_m|\mathbf{x}_n), \overline{P}(y_m|\mathbf{x}_n)\right)$ is the function that generates the numerical representation of the interval

by using the lower and upper bounds, while $h\left(\left|\overline{P.(y_m|\mathbf{x}_n)} - \underline{P.(y_m|\mathbf{x}_n)}\right|\right)$ is a decreasing function of the length of the interval $\left[\underline{P.(y_m|\mathbf{x}_n)}, \overline{P.(y_m|\mathbf{x}_n)}\right]$ which reflects the specificity (or weight) of the numerical value generated by the de-granularization process from $g$.

In this work, the function $g(\cdot)$ is chosen in the form of:

$$g\left(\underline{P.(y_m|\mathbf{x}_n)}, \overline{P.(y_m|\mathbf{x}_n)}\right) = \frac{\underline{P.(y_m|\mathbf{x}_n)} + \overline{P.(y_m|\mathbf{x}_n)}}{2} \tag{10}$$

while $h(\cdot)$ is given by one of these three expressions.

$$h_1\left(\left|\overline{P.(y_m|\mathbf{x}_n)} - \underline{P.(y_m|\mathbf{x}_n)}\right|\right) = 1 \tag{11}$$

$$h_2\left(\left|\overline{P.(y_m|\mathbf{x}_n)} - \underline{P.(y_m|\mathbf{x}_n)}\right|\right) = \frac{1}{\left|\overline{P.(y_m|\mathbf{x}_n)} - \underline{P.(y_m|\mathbf{x}_n)}\right|} \tag{12}$$

$$h_3\left(\left|\overline{P.(y_m|\mathbf{x}_n)} - \underline{P.(y_m|\mathbf{x}_n)}\right|\right) = \exp\left(-\left|\overline{P.(y_m|\mathbf{x}_n)} - \underline{P.(y_m|\mathbf{x}_n)}\right|\right) \tag{13}$$

The Boolean class label of $\mathbf{x}_n$ is then predicted to be in the class with the maximum class membership grades:

$$\mathbf{x}_n \in y_k \text{ if } k = \arg\max_{m=1,\ldots,M} \text{NCM}(\mathbf{x}_n \in y_m) \tag{14}$$

Since $\mathbf{x}_n$ is a training observation, class label of $\mathbf{x}_n$ i.e. $y(\mathbf{x}_n)$ is known in advance. After looping the procedure though all training observations, classification error rate associated with each $\alpha \in \mathcal{A}$ can be computed as:

$$err(\alpha) = \sum_{n=1}^{N} \mathbf{I}[y(\mathbf{x}_n) \neq y_n]/N \tag{15}$$

in which $\mathbf{I}[\Theta] = 1$ if $\Theta$ is true and $0$ if otherwise. The optimal value of $\alpha$ is the one that minimizes $err$. This optimal value will be used as input of the next algorithm to predict the class label for unlabeled observations.

Having value of $\alpha$, the $K$ base classifiers (denoted by $\{BC_k\}_{k=1,\ldots,K}$) are trained by learning $K$ learning algorithms (denoted by $\{\mathcal{K}_k\}_{k=1,\ldots,K}$) on the entire training set.

**Algorithm 2: Training process**

| Input | Training set $\mathcal{D} = \{(\mathbf{x}_n, y(\mathbf{x}_n))\}$, $K$ learning algorithms $\mathcal{K} = \{\mathcal{K}_k | k = 1, \ldots, K\}$, and array $\mathcal{A}$ of values for searching optimal $\alpha$ |
|---|---|
| Output | Optimal value of $\alpha$ and base classifiers $BC_k$ ($k = 1, \ldots, K$) |

(**Generation of meta-data of training set by T-fold CV**)

$\mathbf{L} = \emptyset$, $\mathcal{D} = \mathcal{D}_1 \cup \ldots \cup \mathcal{D}_T$, $\mathcal{D}_i \cap \mathcal{D}_j = \emptyset$ $(i \neq j)$

For each $\mathcal{D}_i$

$\quad \mathcal{D}^{-i} = \mathcal{D} - \mathcal{D}_i$

$\quad$ For each $\mathcal{K}_k$

$\quad\quad$ Classifier $BC_k^{-i} = \text{Learn}(\mathcal{K}_k, \mathcal{D}^{-i})$

$\quad\quad \mathbf{L} = \mathbf{L} \cup \text{Classify}(BC_k^{-i}, \mathcal{D}_i)$ as in (1b)

$\quad$ End For

End For

(**Computing error rate corresponding to each $\alpha$**)

For each $\alpha \in \mathcal{A}$

$\quad$ For $n^{th}$ row of $\mathbf{L}$

$\quad\quad$ Call Algorithm1 on each column of $\mathbf{L}(\mathbf{x}_n)$ with $\alpha$

$\quad\quad$ to obtain interval $\left[\underline{P(y_m|\mathbf{x}_n)}, \overline{P(y_m|\mathbf{x}_n)}\right]$ $(m = 1, \ldots, M)$

$\quad\quad$ Compute $\text{NCM}(\mathbf{x}_n \in y_m)$ $(m = 1, \ldots, M)$ by (9)

$\quad\quad$ Assign class label $y_n$ to $\mathbf{x}_n$ by (14)

```
        End

    Compute  $err(\alpha) = \sum_{n=1}^{N} \mathbf{I}[y(\mathbf{x}_n) \neq y_n]/N$

End
```

**(Choosing optimal $\alpha$)**

$$\alpha_{opt} = \arg\min_{\alpha \in \mathcal{A}} err(\alpha)$$

**(Learning the ensemble base classifiers)**

```
For each  $\mathcal{K}_k$

    $BC_k$ = Learn $(\mathcal{K}_k, \mathcal{D})$

End for

Return  $\alpha_{opt}$  and  $BC_k$   $(k = 1, \ldots, K)$
```

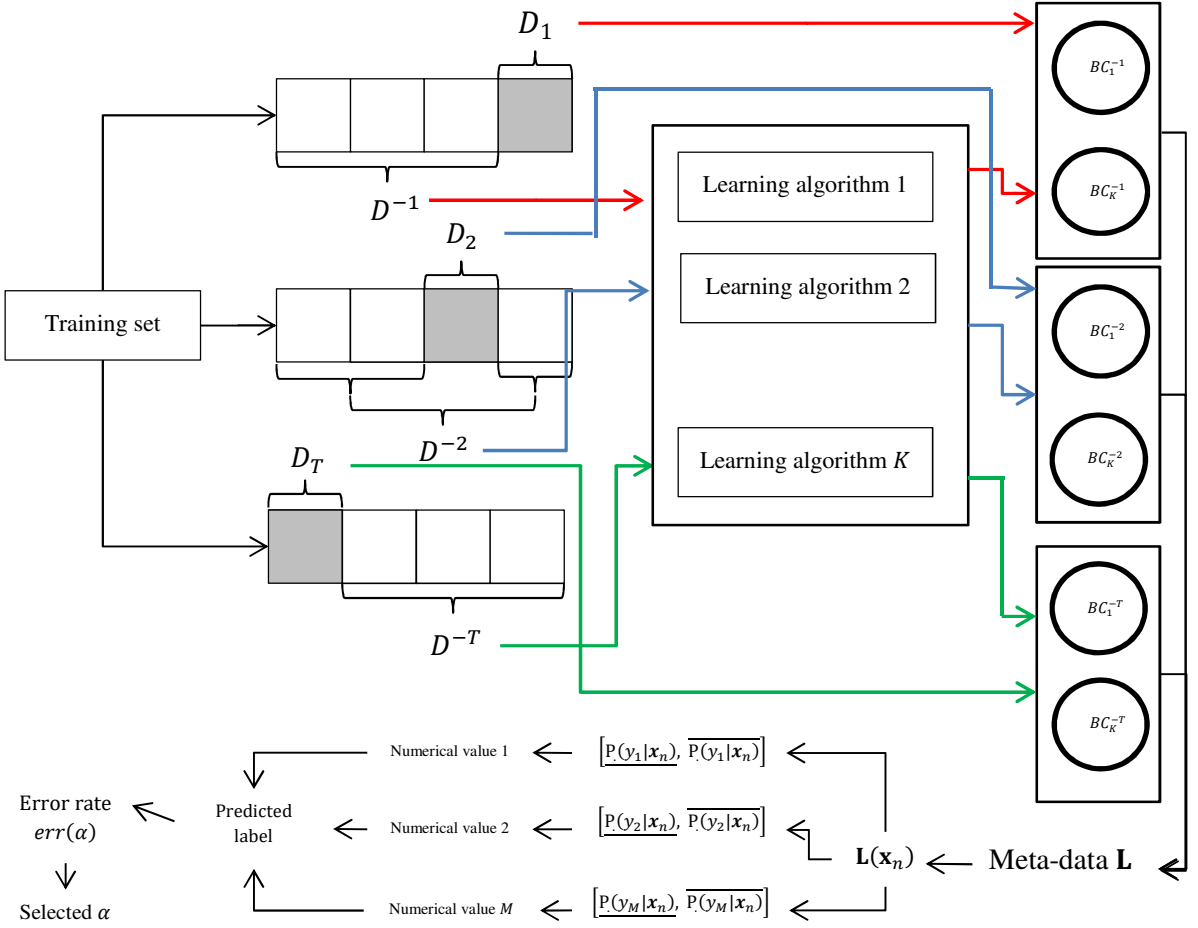

**Fig.1. Generating meta-data and choosing α in Algorithm 1**

In the classification process, for an unlabeled observation $\mathbf{x}^u$, we use the trained base classifiers $\{BC_k\}_{k=1,\ldots,K}$ to obtain the meta-data of $\mathbf{x}^u$ as in (1a). In detail, meta-data of $\mathbf{x}^u$ associated with base classifier $BC_k$ is obtained in the form of vector $\left(P_k(y_1|\mathbf{x}^u),\ldots,P_k(y_M|\mathbf{x}^u)\right)$ in which $P_k(y_m|\mathbf{x}^u)$ is the posterior probability that $\mathbf{x}^u$ belongs to class $y_m$ given by $BC_k$. After that, interval membership values for each class prediction are computed from the meta-data as in (8) i.e. $\left[\underline{P.}(y_m|\mathbf{x}^u),\overline{P.(y_m|\mathbf{x}^u)}\right]$ ($m=1,\ldots,M$). Finally, the classification is obtained by (14). We arrive at the following classification process based on justifiable granularity:

**Algorithm 3: Predicting label for unlabeled observation**

| Input: | Unlabeled observation set $\mathcal{S}=\{\mathbf{x}^u\}$, base classifiers $BC_k$ $(k=1,\dots,K)$, and $\alpha$ |
|---|---|
| Output: | Predicted class label for observations in $\mathcal{S}$ |

---

**(Generating Meta-data)**

For each $\mathbf{x}^u$ in $\mathcal{S}$

    $\mathbf{L}(\mathbf{x}) = \emptyset$

    For each $BC_k$ $(k=1,\dots,K)$

        $\mathbf{L}(\mathbf{x}^u) = \mathbf{L}(\mathbf{x}^u) \cup \text{Classify}(BC_k, \mathbf{x})$

    End

End

---

**(Building intervals and assigning class label)**

For each $\mathbf{x}^u$ in $\mathcal{S}$

    Call **Algorithm 1** with $\alpha$ to find interval $\left[\underline{P.(y_m|\mathbf{x}^u)}, \overline{P.(y_m|\mathbf{x}^u)}\right]$ for $m^{th}$ column of (1a) as $\mathbf{L}_m(\mathbf{x}^u) = \{P_k(y_m|\mathbf{x}^u)\}_{k=1,\dots,K}$ $(m=1,\dots,M)$

    Compute $\text{NCM}(\mathbf{x}^u \in y_m)$ $(m=1,\dots,M)$ by (9)

    Assign class label to $\mathbf{x}^u$ by (14)

End For

---

Clearly, the proposed method described above is a trainable combining method because the meta-data of training observations is exploited to find the value of $\alpha$ in the training process. If a specific value of $\alpha$ is used, the proposed method becomes a fixed combining method in which the label in the meta-data of training set is not used to train the combiner. In the experiment, we evaluate the proposed method in both cases i.e. trainable and fixed combining method.

## 4. Experimental Studies

### 4.1. Datasets and Experimental Settings

To evaluate the performance of the proposed method, we carried out experiments on twenty one UCI datasets as shown in Table 2. These datasets are often used to assess the performance of classification systems [48].

TABLE.2. INFORMATION OF UCI DATASETS IN EVALUATION

| File name | # of features | # of observations | # of classes |
|---|---|---|---|
| Abalone | 8 | 4174 | 3 |
| Artificial | 10 | 700 | 2 |
| Australian | 14 | 690 | 2 |
| Blood | 4 | 748 | 2 |
| Bupa | 6 | 345 | 2 |
| Contraceptive | 9 | 1473 | 3 |
| Dermatology | 34 | 358 | 6 |
| Fertility | 9 | 100 | 2 |
| Haberman | 3 | 306 | 2 |
| Heart | 13 | 270 | 2 |
| Penbased | 16 | 10992 | 10 |
| Pima | 8 | 768 | 2 |
| Plant Margin | 64 | 1600 | 100 |
| Satimage | 36 | 6435 | 6 |
| Skin_NonSkin | 3 | 245057 | 2 |
| Tae | 20 | 151 | 3 |
| Texture | 40 | 5500 | 10 |
| Twonorm | 20 | 7400 | 2 |
| Vehicle | 18 | 946 | 4 |
| Vertebral | 6 | 310 | 3 |
| Yeast | 8 | 1484 | 10 |

We performed extensive comparative studies with a number of existing algorithms as benchmarks: six fixed combining rules, namely Sum, Product, Max, Min, Median, and Majority Vote [34]; one trainable combining methods, namely Decision Template (we used the similarity measure $S_1$ defined as $S_1(\mathbf{L}(\mathbf{x}), DTem_m) = \frac{C\{\mathbf{L}(\mathbf{x}) \cap DTem_m\}}{C\{\mathbf{L}(\mathbf{x}) \cup DTem_m\}}$ where $DTem_m$ is the Decision Template of $m^{th}$ class [7]); three well-known homogeneous ensemble methods, namely AdaBoost [20] (we used Decision

Tree with maximum of 200 iterations as in [19]), Bagging [21], and Random Subspace [23] (we used 200 learners as in [19]).

Ten learning algorithms, namely Linear Discriminant Analysis (denoted by LDA), Naïve Bayes, three K Nearest Neighbor classifiers (with the number of nearest neighbors set to 5, 25, 50, denoted by $KNN_5$, $KNN_{25}$, and $KNN_{50}$, respectively), Decision Tree, Decision Stump, Fisher Classifier [49], Nearest Mean Classifier, and Logistic Linear [50], were chosen to construct the heterogeneous ensemble system. These learning algorithms were chosen to ensure diversity of the ensemble system. The proposed method is compared to the benchmark algorithms with respect to the classification error rate and F1 score (which is the harmonic mean of Precision and Recall) [51]. We performed 10-fold cross validation and run the test 10 times to obtain 100 test results for each dataset. All source codes were implemented in Matlab running on a PC with Intel Core i5 with 2.5 GHz processor and 4G RAM. To assess the statistical significance of the results, i.e., to determine whether the difference in classification error rate is meaningful statistically, we used Wilcoxon signed-rank test [52] (level of significance was set to 0.05) to compare the classification results of our approach and each benchmark algorithm.

### 4.2. Results and Discussion

#### 4.2.1. The influence of $\alpha$ and $h$

We first analyzed the influence of the parameters on the classification results. Here, we evaluated the effect of $\alpha$ on the classification error rate by setting this parameter to one of the values in $\{0, 0.1, 0.2, \ldots, 3.9, 4\}$. For each dataset, we ran the proposed method for each value of $\alpha$, and reported the classification error rate corresponding to the three functions $h_1, h_2$ and $h_3$. The relationships between $\alpha$ and the classification error rate on some datasets are displayed in Fig.2.

Several observations could be made. First, it is interesting to see that the three $h$ functions have very similar error rate profile in the proposed ensemble system on the two-class datasets. Meanwhile, on the other datasets, the error rates related to $h_1$ and $h_3$ are nearly equal and are lower than that of $h_2$. For example, on Contraceptive, Vehicle, Tae, and Yeast, the error rates related to $h_2$ are 3-5% higher than that of $h_1$ and $h_3$. It is noted that $h_2$ is more sensitive to the interval length than the others. Specifically, if the interval length is too small, the function $h_2$ returns large values because $\lim_{x \to 0} \frac{1}{x} = +\infty$. Since some information granule intervals can be very small (see Table A.1), we suggest using $h_1$ or $h_3$ to generate the numerical class memberships from the interval-based information granules. In subsequent discussion, we only report the classification results for $h_3$.

Besides, Fig.2 also shows that the parameter $\alpha$ could have a significant effect on the classification error and their optimum value is somewhat dataset dependent. As shown in Algorithm 2, $\alpha$ can be learned from the given training set. We conducted statistical test to compare the classification error using the specific value of $\alpha = 1$ (called Proposed Specific$_{10}$, where the subscript denotes the number of base classifiers used in the ensemble) versus the optimal value of $\alpha$ (called Proposed CV$_{10}$). The statistical test result showed that using specific value performed worse than using cross validation on five datasets, namely Penbased, Skin&NonSkin, Vehicle, Texture and Tae, while they are equal in performance on the other sixteen datasets. This indicates that in some cases, the specific value of $\alpha = 1$ can be used.

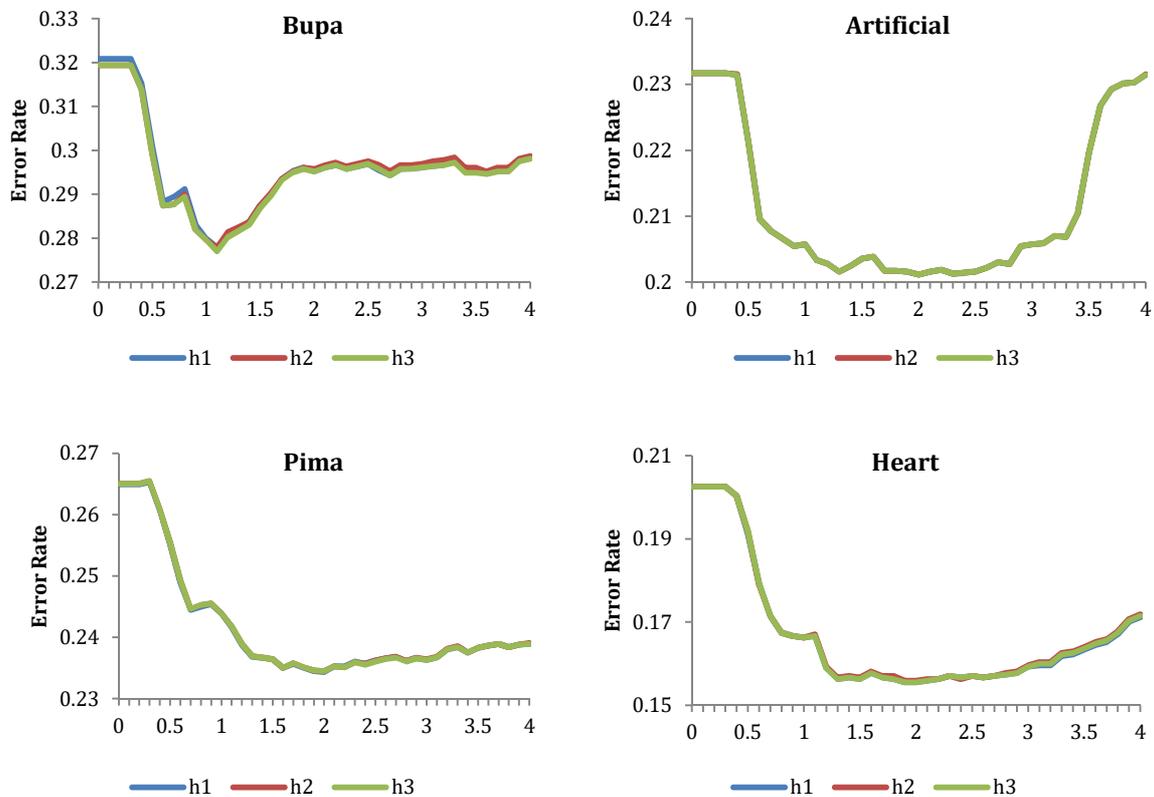

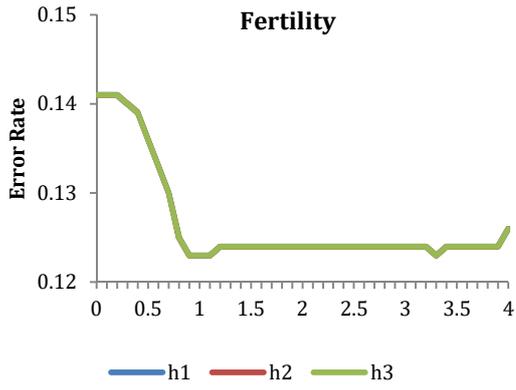
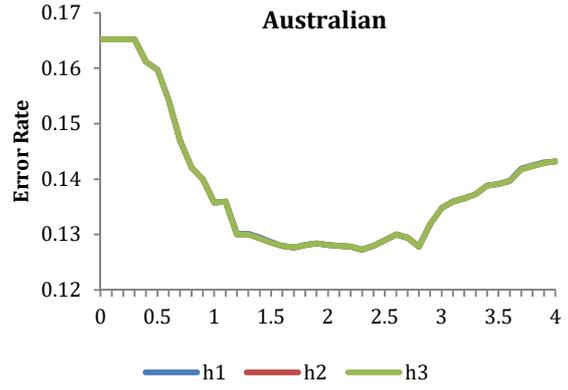
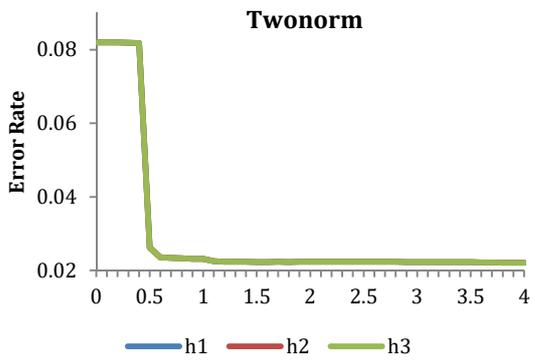
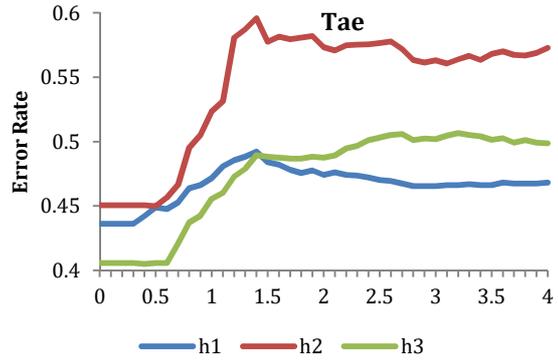
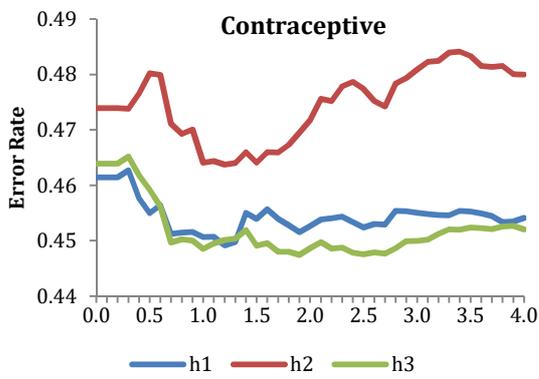
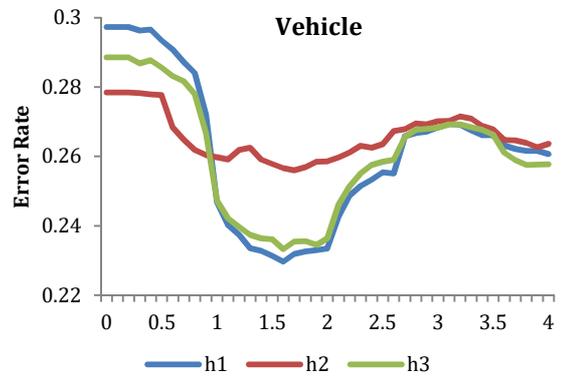
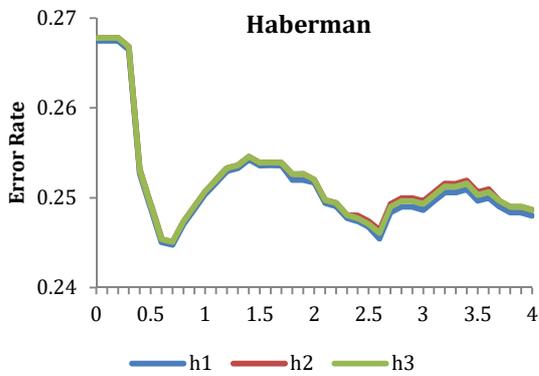
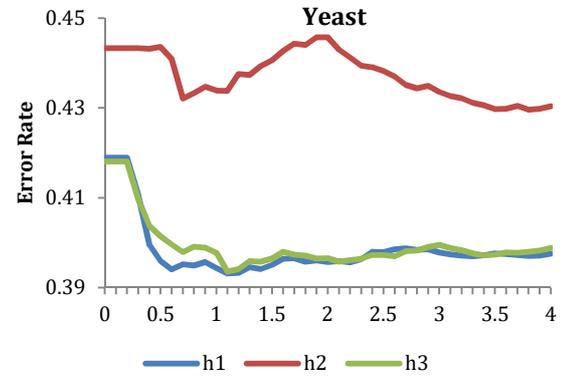

**Fig.2. Effect of parameter $\alpha$ on classification error rate for $h_1, h_2$ and $h_3$ for the twelve selected experimental datasets**

### 4.2.2. Comparison with the benchmark algorithms

The mean and variance of error rates and F1 scores of ten learning algorithms, the benchmark algorithms, and the proposed method (using $h_3$) are reported in Tables A.2 to Table A.7. We first compared the average ranking of the proposed method to the ten learning algorithms [52]. Table 3 shows the average ranking of ten learning algorithms and the proposed methods with respect to the error rate and F1 scores on the experimental datasets. The Proposed $CV_{10}$ and Proposed $Specific_{10}$ are ranked in the first two positions. It demonstrated the benefit of ensemble compared to each of the single learning algorithm.

TABLE.3. AVERAGE RANKINGS OF TEN LEARNING ALGORITHMS AND THE PROPOSED METHOD

| Methods | Ranking based on Error Rate | Ranking based on F1 |
|---|---|---|
| LDA | 5.26 | 5.24 |
| Nave Bayes | 7.81 | 7.14 |
| $KNN_5$ | 6.74 | 6.10 |
| Decision Tree | 7.81 | 6.31 |
| $KNN_{25}$ | 5.5 | 6.05 |
| $KNN_{50}$ | 6.29 | 7.74 |
| Decision Stump | 10.64 | 11.64 |
| Fisher Classifier | 5.95 | 6.98 |
| Logistic Linear | 5.52 | 5.88 |
| Nearest Mean Classier | 10.55 | 8.67 |
| Proposed $CV_{10}$ | 2.19 | 2.38 |
| Proposed $Specific_{10}$ | 3.74 | 3.88 |

TABLE.4. STATISTICAL TEST RESULT COMPARING THE PROPOSED METHOD TO THE BENCHMARK ALGORITHMS (USING TEN LEARNING ALGORITHMS)

| | Proposed $CV_{10}$ | | | | | | Proposed $Specific_{10}$ | | | | | |
|---|---|---|---|---|---|---|---|---|---|---|---|---|
| | ERROR RATE | | | F1 | | | ERROR RATE | | | F1 | | |
| | Win | Equal | Loss | Win | Equal | Loss | Win | Equal | Loss | Win | Equal | Loss |
| Decision Template | 12 | 8 | 1 | 7 | 11 | 3 | 10 | 9 | 2 | 6 | 12 | 3 |
| Sum Rule | 8 | 13 | 0 | 8 | 13 | 0 | 5 | 16 | 0 | 6 | 15 | 0 |
| Product Rule | 17 | 4 | 0 | 17 | 4 | 0 | 17 | 2 | 2 | 15 | 4 | 2 |
| Max Rule | 18 | 3 | 0 | 16 | 5 | 0 | 17 | 3 | 1 | 15 | 5 | 1 |
| Min Rule | 19 | 2 | 0 | 16 | 5 | 0 | 19 | 0 | 2 | 16 | 3 | 2 |
| Median Rule | 11 | 10 | 0 | 11 | 10 | 0 | 6 | 13 | 2 | 7 | 13 | 1 |
| Majority Vote Rule | 11 | 10 | 0 | 13 | 8 | 0 | 9 | 10 | 2 | 12 | 8 | 1 |
| Random Subspace | 16 | 3 | 2 | 16 | 3 | 2 | 15 | 4 | 2 | 13 | 6 | 2 |
| AdaBoost | 18 | 1 | 2 | 16 | 2 | 3 | 17 | 2 | 2 | 14 | 3 | 4 |
| Bagging | 12 | 6 | 3 | 12 | 3 | 6 | 9 | 9 | 3 | 8 | 7 | 6 |

*Level of significance was set to 0.05*

We then compared the Proposed CV$_{10}$ with the benchmark algorithms. The statistical test results displayed in Table 4 show that the proposed method is significantly better than all benchmark algorithms on the experimental datasets. It demonstrates the benefit of using information granules to capture the uncertainty in class label prediction as oppose to just using pointwise information in the meta-data. Note that our framework is not only able to return the numerical class memberships for class label prediction but also the interval membership values that reflect the uncertainty associated with the class prediction by the base classifiers.

In detail, the proposed method with cross validation clearly outperformed all six fixed combining rules. Proposed CV$_{10}$ also outperformed the trainable combining method Decision Template (12 wins vs 1 loss for error rate, and 7 wins vs 3 losses for F1 score). It also achieved better results than the three homogeneous ensemble methods: Bagging (12 wins vs 3 losses for error rate, and 12 wins vs 6 losses for F1 score), Random Subspace (16 wins vs 2 losses for both error rate and F1 score), and Adaboost (18 wins vs 2 losses for error rate, and 16 wins vs 3 losses for F1 score).

When the specific value of $\alpha = 1$ was used, the proposed method is still better than all the fixed combining rules. Proposed Specific$_{10}$ also outperformed Adaboost (17 wins vs 2 losses for error rate, and 14 wins vs 4 losses for F1 score), Bagging (9 wins vs 3 losses for error rate, and 8 wins vs 6 losses for F1 score), Random Subspace (15 wins vs 2 losses for error rate, 13 wins vs 2 losses for F1 score). It also outperformed Decision Template by 10 wins vs 2 losses for error rate and 6 wins vs 3 losses for F1 score.

### 4.2.3. Time complexity analysis

In the case of using a specific value of $\alpha = 1$, the time complexity of training base classifiers is equal to those of other fixed combining counterparts like Sum Rule and Product Rule. Meanwhile, in the case of using optimal value of $\alpha$, the overall time complexity of the proposed method using cross validation will be $\mathcal{O}\left(\max\left(\arg\max_{k=1,\ldots,K} \mathcal{O}(\mathcal{K}_k) \times T, (N \times M \times K \times \log K)\right)\right)$ in which $\mathcal{O}(\arg\max_{k=1,\ldots,K} \mathcal{O}(\mathcal{K}_k) \times T)$ is the time complexity of generating meta-data of training set by running $T$-fold Cross Validation with $\mathcal{K}_k$ learning algorithms ($k = 1, \ldots, K$) having complexity $\mathcal{O}(\mathcal{K}_k)$, and $\mathcal{O}(N \times M \times K \times logK)$ is the time complexity to obtain the interval class memberships for training observations. The time complexity of testing process is $\mathcal{O}(M \times K \times logK)$. Based on the experimental results, our testing process is slightly more complex than other fixed combining methods with longer running time.

### 4.3. Different number of learning algorithms

To demonstrate the effectiveness of the proposed method, five additional learning algorithms, namely Perceptron, Random Neural Network (denoted by RNN, from PRTool5.1 with default values for parameters as in [53]), K Nearest Neighbor classifiers (with the number of nearest neighbors set to 75, denoted by $KNN_{75}$), Discriminative Restricted Boltzmann Machine [54] (denoted by DRBM, from PRTool5.1 with 50 hidden units and default regularization parameter $L_0$ as in [53]), and L2-loss Linear Support Vector Machine [55] (denoted by L2LSVM with default parameter values) were added to the ensemble. (The mean and variance of the classification error rates and F1 scores of the five additional learning algorithms, three homogeneous ensemble methods, seven heterogeneous ensemble methods, and the proposed method with 15 learning algorithms denoted by Proposed $CV_{15}$ and Proposed $Specific_{15}$ can be found in the supplement material). First, the average rankings shown in Table 5 indicated the outstanding performance of the proposed method compared to the 15 learning algorithms, where Proposed $CV_{15}$ ranks first with average ranking of 2.90 and 3.52 for error rate and F1 score, respectively, closely followed by Proposed $Specific_{15}$ (its ranking is 4.33 and 4.55, respectively). Besides, the statistical test results in Table 6 show that both Proposed $CV_{15}$ and Proposed $Specific_{15}$ achieve significantly better performance than all the benchmark algorithms.

TABLE.5. AVERAGE RANKINGS OF FIFTEEN LEARNING ALGORITHMS AND THE PROPOSED METHOD

| Methods | Ranking based on Error Rate | Ranking based on F1 |
|---|---|---|
| LDA | 7.02 | 6.83 |
| Nave Bayes | 10.67 | 9.48 |
| $KNN_5$ | 9.26 | 8.12 |
| Decision Tree | 10.88 | 8.57 |
| $KNN_{25}$ | 7.38 | 8.07 |
| $KNN_{50}$ | 8.33 | 10.12 |
| Decision Stump | 14.95 | 16.52 |
| Fisher Classifier | 7.90 | 9.33 |
| Logistic Linear | 7.10 | 7.83 |
| Nearest Mean Classier | 14.79 | 11.88 |
| L2LSVM | 8.40 | 9.48 |
| Perceptron | 12.50 | 11.52 |
| RNN | 7.57 | 6.88 |
| 8KNN$_{75}$ | 10.21 | 11.74 |
| DRBM | 8.79 | 8.55 |
| Proposed CV$_{15}$ | 2.90 | 3.52 |
| P2roposed Specific$_{15}$ | 4.33 | 4.55 |

TABLE.6. STATISTICAL TEST RESULT COMPARING THE PROPOSED METHOD TO THE BENCHMARK ALGORITHMS (USING FIFTEEN LEARNING ALGORITHMS)

| | Proposed CV$_{15}$ | | | | | | Proposed Specific$_{15}$ | | | | | |
|---|---|---|---|---|---|---|---|---|---|---|---|---|
| | ERROR RATE | | | F1 | | | ERROR RATE | | | F1 | | |
| | Win | Equal | Loss | Win | Equal | Loss | Win | Equal | Loss | Win | Equal | Loss |
| Decision Template | 12 | 8 | 1 | 7 | 10 | 4 | 10 | 10 | 1 | 6 | 11 | 4 |
| Sum Rule | 6 | 14 | 1 | 6 | 14 | 1 | 3 | 17 | 1 | 5 | 15 | 1 |
| Product Rule | 19 | 2 | 0 | 15 | 6 | 0 | 18 | 2 | 1 | 14 | 5 | 2 |
| Max Rule | 19 | 2 | 0 | 15 | 4 | 2 | 19 | 1 | 1 | 13 | 5 | 3 |
| Min Rule | 19 | 2 | 0 | 15 | 4 | 2 | 18 | 2 | 1 | 14 | 4 | 3 |
| Median Rule | 13 | 8 | 0 | 13 | 8 | 0 | 9 | 12 | 0 | 13 | 8 | 0 |
| Majority Vote Rule | 11 | 10 | 0 | 14 | 7 | 0 | 11 | 10 | 0 | 15 | 6 | 0 |
| Random Subspace | 15 | 4 | 2 | 15 | 3 | 3 | 15 | 4 | 2 | 13 | 5 | 3 |
| AdaBoost | 17 | 3 | 1 | 15 | 3 | 3 | 15 | 4 | 2 | 13 | 4 | 4 |
| Bagging | 10 | 8 | 3 | 12 | 4 | 5 | 8 | 9 | 4 | 8 | 6 | 7 |

*Level of significance was set to 0.05

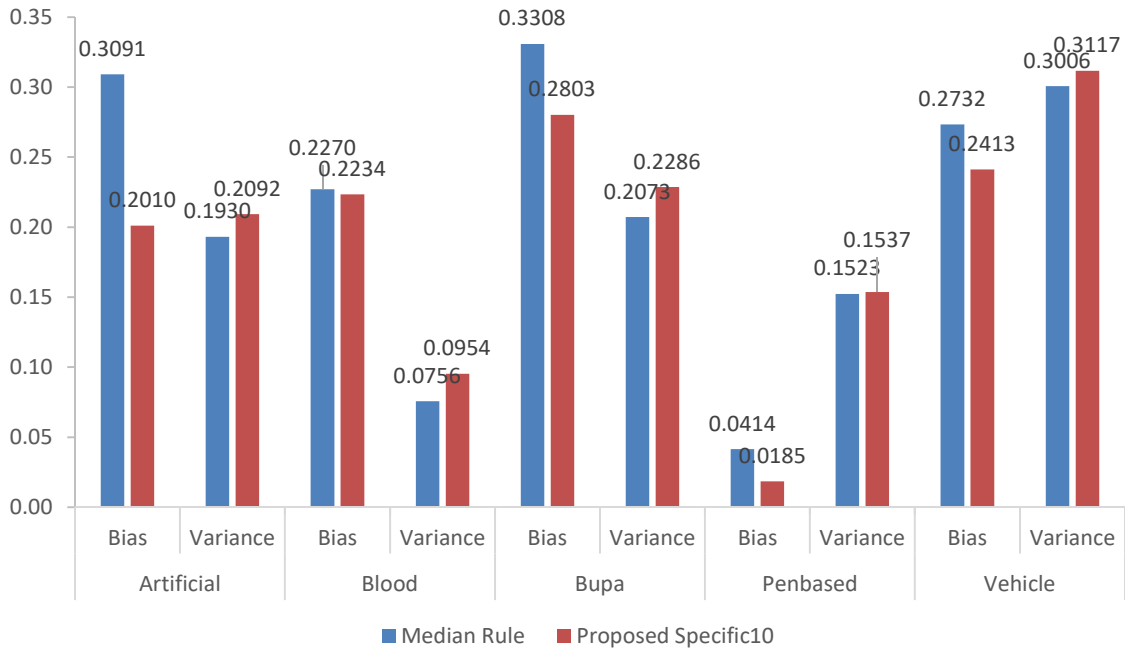

Fig.3. The bias and variance of Proposed Specific$_{10}$ and Median Rule

### 4.4. Bias-variance comparison

Bias-variance theorem is often used to demonstrate that ensemble methods can reduce bias without tradeoff in variance [56]. However we are not aware of any specific study on bias-variance tradeoff for heterogeneous ensemble classification systems. Compare with ensemble systems using pointwise decision model, our interval-based approach based on information granularity offers greater flexibility for the final decision model. In learning theory, a learner with greater flexibility would have higher variance in the bias-variance tradeoff. Here, we compare the Proposed Specific$_{10}$ to Median Rule combining method as both are fixed combining methods consisting of a heterogeneous ensemble of ten learning algorithms. Since in Algorithm 1 the upper and lower bounds of interval are obtained around the median, Median Rule method can be viewed as a special case of the proposed approach where the interval shrinks to a single point. Note that Median Rule obtains the decision model by getting the maximum of the medians (i.e. pointwise information) of the posterior probabilities associated with each class label, Proposed Specific$_{10}$ gets intervals from the posterior probabilities and then forms the decision model by considering both interval's length and bounds. In this paper, we computed *bias* and *variance* based on the 0-1 loss function [56]

$$bias = \frac{\sum_{\mathbf{x} \in S} \mathbf{I}(h(\mathbf{x}) \neq y(\mathbf{x}))}{|S|} \tag{16}$$

$$variance = \frac{\sum_{\mathbf{x} \in S} \sum_{k=1}^{K} \mathbf{I}(h(\mathbf{x}) \neq h_k(\mathbf{x}))}{|S| \times K} \tag{17}$$

in which $S$ is a set including $|S|$ observations, $h(\mathbf{x})$ is the final hypothesis obtained by combining $K$ hypotheses generated by the $K$ base classifiers on $\mathbf{x}$ i.e. $h_k(\mathbf{x})$ ($k = 1, ..., K$), and $y(\mathbf{x})$ is the class label of $\mathbf{x}$.

Fig.3. shows the comparison based on bias and variance on several datasets for Median Rule and Proposed Specific$_{10}$. The values of bias and variance are computed on each of the 100 tests based on running 10-CV procedure 10 times. We see from Fig. 3 that Proposed Specific$_{10}$ has noticeably lower bias and slightly higher variance than Median Rule on those datasets, i.e., on average, there is a reduction of 23.76% in bias and only an increase of 9.89% in variance. Hence, our proposed approach significantly reduces classification bias compared to its non-interval-based counterpart, with only a slight increase in variance due to greater flexibility.

## 5. Conclusions

In this paper, we have introduced a novel fixed combining classifiers ensemble method based on the justifiable granularity concept. Instead of using a single membership value given by pointwise statistics such as the mean, maximum, minimum, or median, we applied the justifiable granularity concept on the meta-data to find the interval associated with each class prediction. This interval reflects the uncertainty in class prediction given by the base classifiers and is a richer representation of information in the meta-data. The numerical class memberships can then be computed from these intervals by considering their bounds and interval length for class label prediction. Extensive experiments were conducted using an ensemble system of ten and fifteen base classifiers, and performance comparison with respect to classification error rate and F1 score was done with several benchmark algorithms on twenty one UCI datasets. The results of statistical testing indicated that our method outperformed the six fixed combining methods, three well-known homogenous ensemble methods, i.e. Adaboost, Bagging, and Random Subspace, and one trainable heterogeneous ensemble method, i.e. Decision Template. The classification accuracy of our proposed ensemble system can be further improved by applying classifier and feature selection on the ensemble system as in [27-30]. Moreover, other designs of information granule such as in [45,46,57] could also be studied. These will be the directions of our future work.

**Acknowledgments**

Tien Thanh Nguyen acknowledges the support of a Griffith University International Postgraduate Research Scholarship (GUIPRS).

**Appendix**

TABLE.A1. INTERVAL CLASS MEMBERSHIPS OF TWO RANDOMLY SELECTED DATA POINTS FROM EACH DATASET

| Datasets | Interval class memberships | | | |
|---|---|---|---|---|
| Bupa | Class1: [0.4066, 0.9032] | Class2: [0.0968, 0.5934] | Class1: [0, 0.4974] | Class2: [0.5026, 1] |
| Artificial | Class1: [0.4770, 1.0000] | Class2:[0, 0.5230] | Class1: [0.4773, 0.9884] | Class2: [0.0116, 0.5227] |
| Pima | Class1: [0.3464, 0.9615] | Class2: [0.0385, 0.6536] | Class1: [0, 0.3464] | Class2: [0.6536, 1] |
| Heart | Class1:[0.5458, 1] | Class2: [0, 0.4542] | Class1: [0.5458, 0.9998] | Class2:[0.0002, 0.4542] |
| Fertility | Class1:[ 0.5373, 1] | Class2:[0, 0.4627] | Class1:[ 0.6, 0.9756] | Class2: [0.0244, 0.4] |
| Australian | Class1: [0.0019, 0.4269] | Class2: [0.5731, 0.9981] | Class1: [0.0033, 0.64] | Class2: [0.36, 0.9967] |
| Twonorm | Class1: [0.995, 1] | Class2:[ 0, 0.005] | Class1: [0, 0] | Class2: [1, 1] |
| Tae | Class1: [0, 0.36] | Class2: [0, 0.34] | Class1: [0.05, 0.4] | Class2: [0, 0.7296] |
| | Class3:[0.3203, 0.8] | | Class3: [ 0.1827, 0.5] | |
| Contraceptive | Class1: [ 0.1114, 0.8571] | Class2: [0.0597, 0.2417] | Class1:[ 0.0551, 0.75] | Class2: [0, 0.2417] |
| | Class3: [0, 0.6] | | Class3:[0.25, 0.6] | |
| Vehicle | Class1: [0, 0.0028] | Class2: [0.1701, 0.9082] | Class1: [0.4320, 1] | Class2: [0, 0.22] |
| | Class3: [0, 0.2308] | Class4: [0.0904, 0.5573] | Class3: [0, 0.0002] | Class4: [0, 0.2] |
| Haberman | Class1:[ 0.6762, 1] | Class2: [0, 0.3238] | Class1: [0.7052, 1] | Class2:[0, 0.2948] |
| Yeast | Class1: [0.4579, 0.8929] | Class2: [0.0714, 0.3599] | Class1: [0.3978, 0.7406] | Class2: [0, 0.4] |
| | Class3:[0, 0.1483] | Class4:[0.0101, 0.2360] | Class3: [0, 0.4] | Class4: [0, 0.0117] |
| | Class5:[0, 0.0638] | Class6:[0, 0.0638] | Class 5:[0, 0.0104] | Class6:[0, 0.0027] |
| | Class7:[0, 0.0319] | Class8:[0, 0.1084] | Class7: [0, 0.0192] | Class8:[0, 0.0188] |
| | Class 9:[ 0,0.0144] | Class10: [0, 0.0048] | Class9: [0, 0.0200] | Class10: [0, 0.0041] |

TABLE.A2. CLASSIFICATION ERROR RATES AND VARIANCES OF TEN LEARNING ALGORITHMS AND THE PROPOSED METHOD

|  | LDA | | Naïve Bayes | | KNN$_5$ | | Decision Tree | | KNN$_{25}$ | | KNN$_{50}$ | |
| --- | --- | --- | --- | --- | --- | --- | --- | --- | --- | --- | --- | --- |
|  | Mean | Variance | Mean | Variance | Mean | Variance | Mean | Variance | Mean | Variance | Mean | Variance |
| Abalone | 0.4561 | 4.30E-04 | 0.4723 | 4.53E-04 | 0.4725 | 4.67E-04 | 0.4927 | 5.63E-04 | 0.4676 | 4.44E-04 | 0.4654 | 5.14E-04 |
| Artificial | 0.4511 | 1.40E-03 | 0.4521 | 1.40E-03 | 0.2496 | 2.40E-03 | 0.2414 | 2.20E-03 | **0.1956** | 2.31E-03 | 0.2006 | 1.78E-03 |
| Australian | 0.1416 | 1.55E-03 | **0.1297** | 1.71E-03 | 0.3457 | 2.11E-03 | 0.1678 | 2.13E-03 | 0.3258 | 1.89E-03 | 0.3284 | 1.90E-03 |
| Blood | 0.2281 | 3.05E-04 | 0.2453 | 1.11E-03 | 0.2341 | 1.56E-03 | 0.2507 | 1.80E-03 | 0.2407 | 4.40E-04 | 0.2382 | 2.15E-05 |
| Bupa | 0.3693 | 8.30E-03 | 0.4264 | 7.60E-03 | 0.3331 | 6.10E-03 | 0.3514 | 6.10E-03 | 0.3211 | 4.42E-03 | 0.3123 | 4.86E-03 |
| Contraceptive | 0.4992 | 1.40E-03 | 0.5324 | 1.42E-03 | 0.4936 | 1.70E-03 | 0.5317 | 1.28E-03 | 0.4571 | 1.38E-03 | 0.4489 | 1.39E-03 |
| Dermatology | 0.0285 | 7.05E-04 | 0.0397 | 9.84E-04 | 0.1138 | 2.63E-03 | 0.0502 | 1.01E-03 | 0.2464 | 3.39E-03 | 0.3394 | 2.46E-03 |
| Fertility | 0.3460 | 2.01E-02 | 0.3770 | 2.08E-02 | 0.1550 | 4.50E-03 | 0.1730 | 7.20E-03 | **0.1200** | 1.60E-03 | **0.1200** | 1.60E-03 |
| Haberman | 0.2510 | 2.02E-03 | 0.2532 | 1.87E-03 | 0.2818 | 4.75E-03 | 0.3048 | 5.27E-03 | **0.2422** | 2.60E-03 | 0.2504 | 1.65E-03 |
| Heart | 0.1593 | 5.30E-03 | 0.1611 | 5.90E-03 | 0.3348 | 5.10E-03 | 0.2381 | 6.70E-03 | 0.3241 | 6.10E-03 | 0.3633 | 5.59E-03 |
| Penbased | 0.1252 | 8.46E-05 | 0.1908 | 9.02E-05 | **0.0074** | 5.44E-06 | 0.0418 | 4.16E-05 | 0.0166 | 1.60E-05 | 0.0246 | 2.27E-05 |
| Pima | 0.2396 | 2.40E-03 | 0.2668 | 2.00E-03 | 0.2864 | 2.30E-03 | 0.2892 | 1.80E-03 | 0.2522 | 2.10E-03 | 0.2683 | 1.66E-03 |
| Plant Margin | 0.1884 | 4.39E-04 | 0.2054 | 7.94E-04 | 0.2378 | 5.89E-04 | 0.5389 | 1.55E-03 | 0.2835 | 7.00E-04 | 0.3474 | 8.62E-04 |
| Satimage | 0.1598 | 1.28E-04 | 0.2126 | 1.76E-04 | **0.0910** | 1.15E-04 | 0.1411 | 1.23E-04 | 0.1067 | 1.10E-04 | 0.1230 | 1.40E-04 |
| Skin_NonSkin | 6.82E-02 | 1.68E-06 | 1.26E-01 | 3.01E-06 | **4.59E-04** | 1.51E-08 | 7.95E-04 | 3.13E-08 | 7.95E-04 | 3.11E-08 | 1.13E-03 | 4.48E-08 |
| Tae | 0.4612 | 1.21E-02 | 0.4505 | 1.22E-02 | 0.5908 | 1.37E-02 | 0.4275 | 1.06E-02 | 0.5676 | 1.67E-02 | 0.6133 | 1.74E-02 |
| Texture | **0.0053** | 7.93E-06 | 0.2470 | 2.68E-04 | 0.0133 | 2.52E-05 | 0.0756 | 1.34E-04 | 0.0274 | 4.40E-05 | 0.0395 | 5.16E-05 |
| Twonorm | 0.0217 | 3.12E-05 | 0.0217 | 3.13E-05 | 0.0312 | 3.96E-05 | 0.0536 | 4.22E-05 | 0.0249 | 3.29E-05 | 0.0233 | 2.56E-05 |
| Vehicle | 0.2186 | 1.39E-03 | 0.5550 | 2.94E-05 | 0.3502 | 2.35E-03 | 0.2932 | 2.13E-03 | 0.3922 | 2.29E-03 | 0.4244 | 1.89E-03 |
| Vertebral | 0.1965 | 3.69E-03 | 0.2565 | 4.59E-03 | 0.1745 | 2.48E-03 | 0.2068 | 3.08E-03 | 0.1671 | 3.03E-03 | 0.1974 | 3.65E-03 |
| Yeast | 0.4098 | 1.09E-03 | 0.4158 | 1.31E-03 | 0.4373 | 1.60E-03 | 0.4642 | 1.86E-03 | 0.4066 | 1.53E-03 | 0.4140 | 1.57E-03 |

|  | Decision Stump | | Fisher Classifier | | Logistic Linear | | Nearest Mean Classifier | | Proposed CV$_{10}$ | | Proposed Specific$_{10}$ | |
| --- | --- | --- | --- | --- | --- | --- | --- | --- | --- | --- | --- | --- |
|  | Mean | Variance | Mean | Variance | Mean | Variance | Mean | Variance | Mean | Variance | Mean | Variance |
| Abalone | 0.6313 | 1.20E-04 | 0.4557 | 4.00E-04 | **0.4469** | 4.42E-04 | 0.5130 | 4.29E-04 | 0.4529 | 5.09E-04 | 0.4572 | 4.54E-04 |
| Artificial | 0.4234 | 1.69E-04 | 0.3091 | 9.61E-04 | 0.3100 | 1.23E-03 | 0.4931 | 1.86E-03 | 0.2016 | 1.76E-03 | 0.2057 | 1.71E-03 |
| Australian | 0.4154 | 6.27E-04 | 0.1417 | 1.62E-03 | 0.1338 | 1.23E-03 | 0.3455 | 1.41E-03 | 0.1328 | 1.39E-03 | 0.1358 | 2.06E-03 |
| Blood | 0.2379 | 1.69E-05 | 0.2276 | 2.86E-04 | 0.2281 | 3.76E-04 | 0.3330 | 2.68E-03 | **0.2234** | 7.29E-04 | 0.2234 | 7.29E-04 |
| Bupa | 0.3988 | 6.56E-04 | 0.3107 | 4.43E-03 | 0.3130 | 4.33E-03 | 0.4448 | 4.41E-03 | **0.2780** | 4.35E-03 | 0.2796 | 4.43E-03 |
| Contraceptive | 0.5730 | 4.74E-06 | 0.5000 | 1.24E-03 | 0.4891 | 1.27E-03 | 0.6271 | 1.01E-03 | **0.4468** | 1.35E-03 | 0.4485 | 1.59E-03 |
| Dermatology | 0.5144 | 1.64E-03 | 0.0235 | 6.67E-04 | 0.0595 | 1.70E-03 | 0.4922 | 7.52E-03 | **0.0224** | 6.11E-04 | 0.0237 | 5.83E-04 |
| Fertility | **0.1200** | 1.60E-03 | 0.1250 | 1.88E-03 | 0.1510 | 3.90E-03 | 0.3760 | 2.92E-02 | 0.1260 | 2.12E-03 | 0.1230 | 1.77E-03 |
| Haberman | 0.2647 | 8.92E-05 | 0.2604 | 1.75E-03 | 0.2556 | 1.58E-03 | 0.3059 | 9.39E-03 | 0.2505 | 1.57E-03 | 0.2507 | 2.77E-03 |
| Heart | 0.4481 | 2.33E-04 | 0.1630 | 4.61E-03 | 0.1652 | 3.44E-03 | 0.3637 | 8.93E-03 | **0.1544** | 3.57E-03 | 0.1630 | 4.57E-03 |
| Penbased | 0.8066 | 8.78E-06 | 0.1357 | 1.09E-04 | 0.0658 | 6.43E-05 | 0.1876 | 1.07E-04 | 0.0145 | 1.22E-05 | 0.0185 | 1.49E-05 |
| Pima | 0.3495 | 4.58E-05 | 0.2278 | 1.51E-03 | **0.2251** | 1.42E-03 | 0.3672 | 2.09E-03 | 0.2373 | 1.75E-03 | 0.2439 | 2.30E-03 |
| Plant Margin | 0.9876 | 7.66E-07 | 0.3661 | 9.37E-04 | 0.5801 | 1.77E-02 | 0.2274 | 7.00E-04 | **0.1879** | 6.62E-04 | 0.1899 | 6.11E-04 |
| Satimage | 0.5975 | 5.93E-05 | 0.2364 | 6.90E-05 | 0.1637 | 9.45E-05 | 0.2229 | 2.00E-04 | 0.1136 | 1.23E-04 | 0.1140 | 1.38E-04 |
| Skin_NonSkin | 2.08E-01 | 1.17E-10 | 7.48E-02 | 1.97E-06 | 8.12E-02 | 2.00E-06 | 1.76E-01 | 4.37E-06 | 4.84E-04 | 1.81E-08 | 6.24E-04 | 2.51E-08 |
| Tae | 0.6575 | 1.24E-03 | 0.4572 | 1.30E-02 | 0.4583 | 1.35E-02 | 0.6629 | 1.66E-02 | **0.4196** | 1.59E-02 | 0.4555 | 1.89E-02 |
| Texture | 0.7737 | 2.03E-04 | 0.0134 | 2.37E-05 | 0.0969 | 3.63E-02 | 0.2419 | 2.71E-04 | 0.0110 | 1.87E-05 | 0.0121 | 2.25E-05 |
| Twonorm | 0.4930 | 9.03E-06 | 0.0220 | 2.44E-05 | 0.0222 | 2.49E-05 | **0.0216** | 2.32E-05 | 0.0225 | 3.00E-05 | 0.0231 | 2.89E-05 |
| Vehicle | 0.6013 | 4.14E-04 | 0.2326 | 1.51E-03 | **0.2097** | 1.72E-03 | 0.6079 | 1.25E-03 | 0.2332 | 1.33E-03 | 0.2472 | 1.84E-03 |
| Vertebral | 0.2310 | 1.37E-03 | 0.2077 | 3.29E-03 | **0.1481** | 2.60E-03 | 0.2423 | 5.03E-03 | 0.1635 | 2.88E-03 | 0.1642 | 3.12E-03 |
| Yeast | 0.6784 | 3.18E-04 | 0.4649 | 1.31E-03 | 0.4157 | 1.18E-03 | 0.4987 | 1.70E-03 | **0.3950** | 1.00E-03 | 0.3977 | 1.66E-03 |

*Bold values indicates the lowest classification error rate

TABLE.A3. CLASSIFICATION ERROR RATES AND VARIANCES OF SEVEN HETEROGENEOUS ENSEMBLE METHODS (USING TEN LEARNING ALGORITHMS)

|  | Decision Template | | Sum Rule | | Product Rule | | Max Rule | | Min Rule | | Median Rule | | Majority Vote Rule | |
| --- | --- | --- | --- | --- | --- | --- | --- | --- | --- | --- | --- | --- | --- | --- |
|  | Mean | Variance | Mean | Variance | Mean | Variance | Mean | Variance | Mean | Variance | Mean | Variance | Mean | Variance |
| Abalone | 0.4742 ■▼ | 4.27E-04 | 0.4579 ■◊ | 4.49E-04 | 0.4699 ■▼ | 4.86E-04 | 0.4813 ■▼ | 4.15E-04 | 0.4742 ■▼ | 5.63E-04 | 0.4594 ■◊ | 5.16E-04 | 0.4566 ■◊ | 5.37E-04 |
| Artificial | 0.2233 ■▼ | 1.53E-03 | 0.2113 ■◊ | 1.70E-03 | 0.2214 ■▼ | 1.79E-03 | 0.2373 ■▼ | 1.73E-03 | 0.2314 ■▼ | 2.24E-03 | 0.3076 ■▼ | 1.30E-03 | 0.3086 ■▼ | 1.07E-03 |
| Australian | 0.1274 ■◊ | 1.50E-03 | 0.1317 ■◊ | 1.33E-03 | 0.1643 ■▼ | 1.82E-03 | 0.1629 ■▼ | 1.84E-03 | 0.1654 ■▼ | 1.88E-03 | 0.1423 ■◊ | 1.40E-03 | 0.1407 ■◊ | 1.52E-03 |
| Blood | 0.2656 ■▼ | 2.43E-03 | 0.2295 ■▼ | 3.43E-04 | 0.2367 ■ | 1.02E-03 | 0.2373 ■▼ | 1.05E-03 | 0.2373 ■▼ | 1.05E-03 | 0.2296 ■▼ | 3.26E-04 | 0.2271 ■▼ | 3.57E-04 |
| Bupa | 0.3110 ■▼ | 4.35E-03 | 0.2978 ■▼ | 4.25E-03 | 0.3178 ■▼ | 5.46E-03 | 0.3240 ■▼ | 6.79E-03 | 0.3166 ■▼ | 4.96E-03 | 0.3336 ■▼ | 4.73E-03 | 0.3060 ■▼ | 4.04E-03 |
| Contraceptive | 0.4560 ■◊ | 1.69E-03 | 0.4395 ■◊ | 1.60E-03 | 0.4524 ■◊ | 1.32E-03 | 0.4706 ■▼ | 1.65E-03 | 0.4656 ■▼ | 1.47E-03 | 0.4524 ■◊ | 9.76E-04 | 0.4624 ■▼ | 1.28E-03 |
| Dermatology | 0.0229 ■◊ | 6.30E-04 | 0.0243 ■◊ | 6.20E-04 | 0.0575 ■▼ | 1.09E-03 | 0.0525 ■▼ | 1.25E-03 | 0.0629 ■▼ | 1.12E-03 | 0.0260 ■◊ | 6.14E-04 | 0.0210 ■▼ | 4.93E-04 |
| Fertility | 0.3330 ■▼ | 2.30E-02 | 0.1230 ■◊ | 1.97E-03 | 0.1350 ■▼ | 2.88E-03 | 0.1320 ■◊ | 3.38E-03 | 0.1380 ■▼ | 3.16E-03 | 0.1220 ■◊ | 1.92E-03 | 0.1200 ■▼ | 1.60E-03 |
| Haberman | 0.2690 ■▼ | 3.36E-03 | 0.2481 ■◊ | 2.30E-03 | 0.2584 ■▼ | 3.58E-03 | 0.2610 ■◊ | 3.49E-03 | 0.2663 ■▼ | 3.57E-03 | 0.2541 ■◊ | 1.71E-03 | 0.2565 ■▼ | 1.85E-03 |
| Heart | 0.1559 ■◊ | 5.39E-03 | 0.1607 ■◊ | 4.29E-03 | 0.1970 ■▼ | 6.22E-03 | 0.2041 ■▼ | 6.19E-03 | 0.2089 ■▼ | 4.62E-03 | 0.1778 ■▼ | 4.36E-03 | 0.2085 ■▼ | 4.41E-03 |
| Penbased | 0.0239 ■▼ | 1.85E-05 | 0.0308 ■▼ | 2.23E-05 | 0.0269 ■▼ | 2.52E-05 | 0.0197 ■▼ | 1.61E-05 | 0.0271 ■▼ | 2.55E-05 | 0.0414 ■▼ | 3.54E-05 | 0.0397 ■▼ | 2.96E-05 |
| Pima | 0.2432 ■◊ | 1.66E-03 | 0.2366 ■◊ | 1.87E-03 | 0.2634 ■▼ | 2.16E-03 | 0.2676 ■▼ | 1.98E-03 | 0.2607 ■▼ | 2.65E-03 | 0.2316 ■▲ | 1.96E-03 | 0.2325 ■▲ | 1.80E-03 |

| | | | | | | | | | | | | | |
|---|---|---|---|---|---|---|---|---|---|---|---|---|---|
| Plant Margin | 0.1876 ◘ ◊ | 5.81E-04 | 0.1887 ◘ ◊ | 5.76E-04 | 0.4706 ■▼ | 1.34E-03 | 0.3183 ■▼ | 7.88E-04 | 0.4724 ■▼ | 1.41E-03 | 0.1916 ◘ ◊ | 5.99E-04 | 0.2048 ■▼ | 7.31E-04 |
| Satimage | 0.2912 ■▼ | 8.64E-05 | 0.1226 ■▼ | 1.56E-04 | 0.4395 ■▼ | 1.18E-04 | 0.1183 ■▼ | 1.54E-04 | 0.4395 ■▼ | 1.18E-04 | 0.1151 ◘ ◊ | 1.48E-04 | 0.1151 ◘ ◊ | 1.33E-04 |
| Skin_NonSkin | 9.82E-04 ■▼ | 4.45E-08 | 4.75E-03 ■▼ | 2.17E-07 | 5.44E-04 ■▲ | 2.08E-08 | 5.10E-04 ■▲ | 1.96E-08 | 5.10E-04 ■▲ | 1.96E-08 | 3.07E-02 ■▼ | 1.05E-06 | 3.95E-02 ■▼ | 1.37E-06 |
| Tae | 0.4465 ◘ ◊ | 1.18E-02 | 0.4427 ◘ ◊ | 1.78E-02 | 0.4166 ◘ ◊ | 1.42E-02 | 0.4516 ◘ ◊ | 1.25E-02 | 0.4176 ◘ ◊ | 1.65E-02 | 0.4495 ◘ ◊ | 1.29E-02 | 0.4589 ◘ ◊ | 1.17E-02 |
| Texture | 0.0974 ■▼ | 1.01E-05 | 0.0120 ◘ ◊ | 1.83E-05 | 0.0532 ■▼ | 2.81E-04 | 0.0310 ■▼ | 5.94E-05 | 0.0531 ■▼ | 2.83E-04 | 0.0127 ◘ ◊ | 2.12E-05 | 0.0147 ■▼ | 2.35E-05 |
| Twonorm | 0.0219 ■▲ | 2.29E-05 | 0.0222 ◘ ◊ | 2.34E-05 | 0.0740 ■▼ | 8.56E-05 | 0.0791 ■▼ | 9.82E-05 | 0.0801 ■▼ | 1.09E-04 | 0.0218 ■▲ | 2.85E-05 | 0.0217 ■▲ | 2.41E-05 |
| Vehicle | 0.2164 ◘▲ | 1.06E-03 | 0.2444 ◘ ◊ | 1.63E-03 | 0.2711 ■▼ | 2.00E-03 | 0.3027 ■▼ | 1.66E-03 | 0.2815 ■▼ | 1.61E-03 | 0.2771 ■▼ | 1.69E-03 | 0.2950 ■▼ | 2.29E-03 |
| Vertebral | 0.1532 ◘ ◊ | 3.34E-03 | 0.1568 ◘ ◊ | 2.75E-03 | 0.1810 ■▼ | 2.54E-03 | 0.2045 ■▼ | 2.75E-03 | 0.2000 ◘ ◊ | 3.29E-03 | 0.1603 ◘ ◊ | 3.42E-03 | 0.1529 ◘ ◊ | 2.95E-03 |
| Yeast | 0.4056 ■ ◊ | 1.44E-03 | 0.3915 ◘ ◊ | 1.13E-03 | 0.4195 ■▼ | 1.03E-03 | 0.4243 ■▼ | 1.48E-03 | 0.4196 ■▼ | 1.65E-03 | 0.3978 ◘ ◊ | 1.17E-03 | 0.4085 ■▼ | 1.28E-03 |

◘ : *The benchmark algorithm is equal to Proposed $CV_{10}$*, □: *The benchmark algorithm is better than Proposed $CV_{10}$*, ■: *The benchmark algorithm is worse than Proposed $CV_{10}$*

◊: *The benchmark algorithm is equal to Proposed $Specific_{10}$*, ▲: *The benchmark algorithm is better than Proposed $Specific_{10}$*, ▼: *The benchmark algorithm is worse than Proposed $Specific_{10}$*

## TABLE.A4. CLASSIFICATION ERROR RATES AND VARIANCES OF THREE HOMOGENEOUS ENSEMBLE METHODS AND THE PROPOSED METHOD

| | Random Subspace | | AdaBoost | | Bagging | | Proposed $CV_{10}$ | | Proposed $Specific_{10}$ | |
|---|---|---|---|---|---|---|---|---|---|---|
| | Mean | Variance | Mean | Variance | Mean | Variance | Mean | Variance | Mean | Variance |
| Abalone | 0.4676 ■▼ | 5.32E-04 | 0.4672 ■▼ | 3.25E-04 | 0.4522 ◘ ◊ | 5.41E-04 | 0.4529 | 5.09E-04 | 0.4572 ◘ | 4.54E-04 |
| Artificial | 0.2700 ■▼ | 2.03E-03 | 0.2197 ■▼ | 1.90E-03 | 0.2069 ◘ ◊ | 2.30E-03 | 0.2016 | 1.76E-03 | 0.2057 ◘ | 1.71E-03 |
| Australian | 0.1903 ■▼ | 1.65E-03 | 0.1425 ■▼ | 1.53E-03 | 0.1351 ◘ ◊ | 1.68E-03 | 0.1328 | 1.39E-03 | 0.1358 ◘ | 2.06E-03 |
| Blood | 0.2269 ◘ ◊ | 7.86E-04 | 0.2060 □▲ | 8.41E-04 | 0.2322 ■▼ | 1.08E-03 | 0.2234 | 7.29E-04 | 0.2234 ◘ | 7.29E-04 |
| Bupa | 0.3544 ■▼ | 4.90E-03 | 0.2587 □▲ | 3.30E-03 | 0.2741 ◘ ◊ | 4.37E-03 | 0.2780 | 4.35E-03 | 0.2796 ◘ | 4.43E-03 |
| Contraceptive | 0.5730 ■▼ | 4.74E-06 | 0.4996 ■▼ | 8.99E-04 | 0.4627 ■▼ | 1.55E-03 | 0.4468 | 1.35E-03 | 0.4485 ◘ | 1.59E-03 |
| Dermatology | 0.0251 ◘ ◊ | 5.40E-04 | 0.0405 ■▼ | 1.04E-03 | 0.0369 ■▼ | 9.82E-04 | 0.0224 | 6.11E-04 | 0.0237 ◘ | 5.83E-04 |
| Fertility | 0.1320 ◘ ◊ | 2.38E-03 | 0.1600 ■▼ | 9.00E-03 | 0.1260 ◘ ◊ | 4.92E-03 | 0.1260 | 2.12E-03 | 0.1230 ◘ | 1.77E-03 |
| Haberman | 0.2994 ■▼ | 3.78E-03 | 0.2728 ■▼ | 3.85E-03 | 0.3179 ■▼ | 4.93E-03 | 0.2505 | 1.57E-03 | 0.2507 ◘ | 2.77E-03 |
| Heart | 0.2667 ■▼ | 3.98E-03 | 0.1896 ■▼ | 4.67E-03 | 0.1700 ◘ ◊ | 4.80E-03 | 0.1544 | 3.57E-03 | 0.1630 ◘ | 4.57E-03 |
| Penbased | 0.0225 ■▼ | 2.12E-05 | 0.4526 ■▼ | 1.35E-04 | 0.0179 ◘ ◊ | 1.71E-05 | 0.0145 | 1.22E-05 | 0.0185 ■ | 1.49E-05 |
| Pima | 0.2689 ■▼ | 1.89E-03 | 0.2444 ◘ ◊ | 1.97E-03 | 0.2357 ◘ ◊ | 2.04E-03 | 0.2373 | 1.75E-03 | 0.2439 ◘ | 2.30E-03 |
| Plant Margin | 0.1662 □▲ | 6.31E-04 | 0.9556 ■▼ | 6.91E-05 | 0.1957 ■▼ | 7.57E-04 | 0.1879 | 6.62E-04 | 0.1899 ◘ | 6.11E-04 |
| Satimage | 0.0906 ■▼ | 9.27E-05 | 0.2036 ■▼ | 1.46E-04 | 9.11E-02 ◘▲ | 1.02E-04 | 0.1136 | 1.23E-04 | 0.1140 ◘ | 1.38E-04 |
| Skin_NonSkin | 2.62E-03 ■▼ | 8.41E-08 | 4.29E-02 ■▼ | 1.81E-06 | 6.06E-04 ◘ ◊ | 2.19E-08 | 4.84E-04 | 1.81E-08 | 6.24E-04 ◘ | 2.51E-08 |
| Tae | 0.4518 ◘ ◊ | 1.27E-02 | 0.5140 ■▼ | 1.80E-02 | 0.3350 ◘▲ | 1.58E-02 | 0.4196 | 1.59E-02 | 0.4555 ■ | 1.89E-02 |
| Texture | 0.0245 ■▼ | 3.35E-05 | 0.3954 ■▼ | 2.11E-04 | 0.0375 ■▼ | 6.09E-05 | 0.0110 | 1.87E-05 | 0.0121 ■ | 2.25E-05 |
| Twonorm | 0.0292 ■▼ | 3.35E-05 | 0.0310 ■▼ | 3.76E-05 | 0.0273 ■▼ | 3.20E-05 | 0.0225 | 3.00E-05 | 0.0231 ■ | 2.89E-05 |
| Vehicle | 0.2994 ■▼ | 1.97E-03 | 0.4451 ■▼ | 2.87E-03 | 0.2499 ■ ◊ | 1.58E-03 | 0.2332 | 1.33E-03 | 0.2472 ■ | 1.84E-03 |
| Vertebral | 0.2619 ■▼ | 4.08E-03 | 0.2258 ■▼ | 9.57E-04 | 0.1781 ■▼ | 2.36E-03 | 0.1635 | 2.88E-03 | 0.1642 ◘ | 3.12E-03 |
| Yeast | 0.4779 ■▼ | 1.45E-03 | 0.5898 ■▼ | 4.63E-03 | 0.3861 ◘▲ | 1.33E-03 | 0.3950 | 1.00E-03 | 0.3977 ◘ | 1.66E-03 |

◘ : *The benchmark algorithm is equal to Proposed $CV_{10}$*, □: *The benchmark algorithm is better than Proposed $CV_{10}$*, ■: *The benchmark algorithm is worse than Proposed $CV_{10}$*

◊: *The benchmark algorithm is equal to Proposed $Specific_{10}$*, ▲: *The benchmark algorithm is better than Proposed $Specific_{10}$*, ▼: *The benchmark algorithm is worse than Proposed $Specific_{10}$*

## TABLE.A5. F1 SCORES AND VARIANCES OF TEN LEARNING ALGORITHMS AND THE PROPOSED METHOD

| | LDA | | Naïve Bayes | | $KNN_5$ | | Decision Tree | | $KNN_{25}$ | | $KNN_{50}$ | |
|---|---|---|---|---|---|---|---|---|---|---|---|---|
| | Mean | Variance | Mean | Variance | Mean | Variance | Mean | Variance | Mean | Variance | Mean | Variance |
| Abalone | 0.5344 | 4.39E-04 | 0.5214 | 4.29E-04 | 0.5277 | 4.59E-04 | 0.5096 | 5.39E-04 | 0.5290 | 4.43E-04 | 0.5296 | 5.10E-04 |
| Artificial | 0.6063 | 3.30E-03 | 0.6060 | 3.24E-03 | 0.7522 | 2.49E-03 | 0.7572 | 2.98E-03 | **0.7979** | 2.42E-03 | 0.7822 | 2.71E-03 |
| Australian | 0.8553 | 1.41E-03 | **0.8703** | 1.39E-03 | 0.6432 | 3.20E-03 | 0.8409 | 1.43E-03 | 0.6389 | 3.36E-03 | 0.6248 | 2.24E-03 |
| Blood | 0.5159 | 3.55E-03 | 0.5633 | 3.40E-03 | 0.6069 | 4.23E-03 | **0.6131** | 4.26E-03 | 0.4834 | 2.06E-03 | 0.4324 | 2.22E-06 |
| Bupa | 0.6575 | 5.46E-03 | 0.5437 | 5.15E-03 | 0.6375 | 6.27E-03 | 0.6352 | 6.74E-03 | 0.6387 | 6.85E-03 | 0.6497 | 6.88E-03 |
| Contraceptive | 0.4930 | 1.70E-03 | 0.4639 | 2.00E-03 | 0.4815 | 1.46E-03 | 0.4944 | 1.78E-03 | 0.5236 | 1.47E-03 | 0.5280 | 1.51E-03 |
| Dermatology | 0.9679 | 9.24E-04 | 0.9514 | 1.73E-03 | 0.8648 | 3.62E-03 | 0.9338 | 2.10E-03 | 0.6925 | 5.65E-03 | 0.5513 | 4.09E-03 |
| Fertility | 0.4544 | 7.49E-04 | 0.4617 | 3.97E-04 | 0.4583 | 8.95E-04 | **0.5217** | 1.96E-02 | 0.4678 | 1.37E-04 | 0.4678 | 1.37E-04 |
| Haberman | 0.5496 | 1.05E-02 | 0.5228 | 1.16E-02 | 0.5737 | 7.69E-03 | 0.5547 | 1.18E-02 | 0.5747 | 1.08E-02 | 0.5206 | 7.17E-03 |
| Heart | 0.8319 | 4.58E-03 | 0.8346 | 4.92E-03 | 0.6714 | 6.69E-03 | 0.7577 | 8.31E-03 | 0.6730 | 8.74E-03 | 0.6121 | 8.03E-03 |
| Penbased | 0.8728 | 8.81E-05 | 0.7976 | 1.05E-04 | **0.9926** | 5.40E-06 | 0.9583 | 4.14E-05 | 0.9835 | 1.60E-05 | 0.9754 | 2.29E-05 |
| Pima | 0.7307 | 3.35E-03 | 0.7118 | 2.86E-03 | 0.6698 | 2.69E-03 | 0.6857 | 3.35E-03 | 0.6924 | 3.33E-03 | 0.6692 | 3.74E-03 |
| Plant Margin | 0.7862 | 7.11E-04 | 0.7667 | 1.26E-03 | 0.7344 | 8.07E-04 | 0.4190 | 1.59E-03 | 0.6777 | 8.96E-04 | 0.6041 | 1.07E-03 |
| Satimage | 0.7926 | 2.25E-04 | 0.7689 | 2.09E-04 | **0.8935** | 1.64E-04 | 0.8356 | 1.69E-04 | 0.8731 | 1.71E-04 | 0.8523 | 2.22E-04 |
| Skin_NonSkin | 0.9013 | 3.28E-06 | 0.8013 | 7.71E-06 | **0.9993** | 3.48E-08 | 0.9988 | 7.22E-08 | 0.9988 | 7.14E-08 | 0.9983 | 1.03E-07 |
| Tae | 0.5034 | 1.57E-02 | 0.5132 | 1.46E-02 | 0.3943 | 1.51E-02 | **0.5499** | 1.21E-02 | 0.4097 | 1.46E-02 | 0.3456 | 1.71E-02 |
| Texture | **0.9943** | 9.55E-06 | 0.7376 | 3.01E-04 | 0.9856 | 3.00E-05 | 0.9197 | 1.47E-04 | 0.9702 | 5.24E-05 | 0.9572 | 5.99E-05 |
| Twonorm | **0.9777** | 2.96E-05 | 0.9781 | 3.15E-05 | 0.9683 | 3.84E-05 | 0.8410 | 1.85E-04 | 0.9746 | 4.44E-05 | 0.9766 | 3.26E-05 |
| Vehicle | 0.7799 | 1.55E-03 | 0.4370 | 2.90E-03 | 0.6395 | 2.05E-03 | 0.7096 | 1.83E-03 | 0.5887 | 2.28E-03 | 0.5375 | 2.15E-03 |
| Vertebral | 0.7687 | 5.27E-03 | 0.7107 | 6.23E-03 | 0.7792 | 4.19E-03 | 0.7288 | 5.47E-03 | 0.7758 | 5.59E-03 | 0.7239 | 7.28E-03 |
| Yeast | 0.5107 | 3.99E-03 | 0.4997 | 3.76E-03 | 0.4956 | 4.45E-03 | 0.4175 | 2.73E-03 | 0.4488 | 2.94E-03 | 0.3907 | 1.18E-03 |

| | Decision Stump | | Fisher Classifier | | Logistic Linear | | Nearest Mean Classifier | | Proposed $CV_{10}$ | | Proposed $Specific_{10}$ | |
|---|---|---|---|---|---|---|---|---|---|---|---|---|
| | Mean | Variance | Mean | Variance | Mean | Variance | Mean | Variance | Mean | Variance | Mean | Variance |
| Abalone | 0.1902 | 7.73E-04 | 0.5255 | 4.03E-04 | **0.5403** | 4.67E-04 | 0.4611 | 4.51E-04 | 0.5373 | 5.21E-04 | 0.5364 | 4.76E-04 |
| Artificial | 0.3781 | 7.48E-04 | 0.6063 | 3.30E-03 | 0.6095 | 3.18E-03 | 0.4725 | 2.84E-03 | 0.7857 | 2.59E-03 | 0.7776 | 2.48E-03 |
| Australian | 0.4329 | 1.68E-03 | 0.8553 | 1.41E-03 | 0.8600 | 1.24E-03 | 0.5980 | 3.61E-03 | 0.8649 | 1.45E-03 | 0.8583 | 1.48E-03 |
| Blood | 0.4325 | 1.74E-06 | 0.5037 | 2.78E-03 | 0.5326 | 3.06E-03 | 0.5712 | 4.02E-03 | 0.5822 | 3.78E-03 | 0.5728 | 3.73E-03 |
| Bupa | 0.4103 | 2.31E-03 | 0.6566 | 5.60E-03 | 0.6626 | 5.65E-03 | 0.5422 | 8.00E-03 | **0.6903** | 5.58E-03 | 0.6872 | 5.57E-03 |
| Contraceptive | 0.1995 | 5.10E-07 | 0.4447 | 1.38E-03 | 0.4815 | 1.52E-03 | 0.3459 | 9.03E-04 | **0.5342** | 1.70E-03 | 0.5332 | 2.05E-03 |
| Dermatology | 0.2145 | 3.34E-04 | 0.9719 | 1.07E-03 | 0.9344 | 2.33E-03 | 0.4644 | 8.15E-03 | **0.9754** | 7.44E-04 | 0.9737 | 7.27E-04 |
| Fertility | 0.4678 | 1.37E-04 | 0.4651 | 2.17E-04 | 0.4556 | 1.14E-03 | 0.4728 | 2.16E-02 | 0.4743 | 4.67E-03 | 0.4663 | 1.81E-04 |
| Haberman | 0.4237 | 1.00E-05 | 0.5223 | 8.25E-03 | 0.5364 | 9.09E-03 | **0.6077** | 9.14E-03 | 0.5662 | 1.08E-02 | 0.5438 | 8.95E-03 |
| Heart | 0.3681 | 1.14E-03 | 0.8319 | 4.58E-03 | 0.8280 | 4.15E-03 | 0.6248 | 7.88E-03 | **0.8405** | 4.56E-03 | 0.8281 | 5.70E-03 |
| Penbased | 0.0648 | 5.45E-07 | 0.8628 | 1.12E-04 | 0.9337 | 6.53E-05 | 0.8059 | 1.19E-04 | 0.9856 | 1.21E-05 | 0.9815 | 1.48E-05 |
| Pima | 0.3997 | 2.56E-04 | 0.7294 | 3.34E-03 | **0.7348** | 3.14E-03 | 0.5916 | 2.62E-03 | 0.7189 | 3.28E-03 | 0.7170 | 3.07E-03 |
| Plant Margin | 0.0005 | 1.27E-09 | 0.5882 | 1.15E-03 | 0.3958 | 1.44E-02 | 0.7435 | 8.42E-04 | **0.7878** | 1.00E-03 | 0.7854 | 8.28E-04 |
| Satimage | 0.1870 | 1.43E-05 | 0.6078 | 1.24E-04 | 0.7586 | 1.84E-04 | 0.7633 | 2.17E-04 | 0.8637 | 1.81E-04 | 0.8657 | 1.90E-04 |
| Skin_NonSkin | 0.4421 | 1.13E-11 | 0.8903 | 4.10E-06 | 0.8783 | 4.54E-06 | 0.7751 | 5.78E-06 | 0.9993 | 4.17E-08 | 0.9991 | 5.78E-08 |
| Tae | 0.1953 | 2.17E-03 | 0.5083 | 1.46E-02 | 0.5080 | 1.50E-02 | 0.2947 | 1.06E-02 | 0.5492 | 1.58E-02 | 0.5219 | 1.69E-02 |
| Texture | 0.0705 | 1.76E-05 | 0.9853 | 2.85E-05 | 0.9053 | 3.37E-02 | 0.7409 | 3.03E-04 | 0.9880 | 2.18E-05 | 0.9868 | 2.64E-05 |
| Twonorm | 0.3507 | 4.01E-05 | 0.9777 | 2.96E-05 | 0.9777 | 2.95E-05 | 0.9781 | 3.15E-05 | 0.9774 | 3.15E-05 | 0.9766 | 3.38E-05 |
| Vehicle | 0.2646 | 2.07E-04 | 0.7631 | 1.75E-03 | **0.7938** | 1.61E-03 | 0.3199 | 1.03E-03 | 0.7697 | 1.80E-03 | 0.7478 | 1.73E-03 |
| Vertebral | 0.5626 | 7.35E-04 | 0.7304 | 5.95E-03 | **0.8067** | 4.51E-03 | 0.7276 | 5.84E-03 | 0.7935 | 5.30E-03 | 0.7863 | 5.83E-03 |
| Yeast | 0.0840 | 1.68E-04 | 0.2946 | 1.19E-03 | 0.4929 | 5.99E-03 | 0.4748 | 3.88E-03 | **0.5254** | 4.34E-03 | 0.5221 | 4.33E-03 |

*Bold value indicates the lowest F1 score

## TABLE.A6. F1 SCORES AND VARIANCES OF SEVEN HETEROGENEOUS ENSEMBLE METHODS (USING TEN LEARNING ALGORITHMS)

| | Decision Template 10 | | Sum Rule | | Product Rule | | Max Rule | | Min Rule | | Median Rule | | Majority Vote Rule | |
|---|---|---|---|---|---|---|---|---|---|---|---|---|---|---|
| | Mean | Variance | Mean | Variance | Mean | Variance | Mean | Variance | Mean | Variance | Mean | Variance | Mean | Variance |
| Abalone | 0.5024 ■▼ | 4.45E-04 | 0.5356 ◘ ◊ | 4.58E-04 | 0.5298 ◘ ▼ | 4.75E-04 | 0.5088 ■▼ | 4.37E-04 | 0.5257 ■▼ | 5.59E-04 | 0.5291 ■▼ | 5.47E-04 | 0.5339 ◘ ◊ | 5.48E-04 |
| Artificial | 0.7729 ◘ ◊ | 3.34E-03 | 0.7810 ■▼ | 2.41E-03 | 0.7670 ◘ ◊ | 3.30E-03 | 0.7610 ■▼ | 3.16E-03 | 0.7610 ■▼ | 3.16E-03 | 0.6120 ■▼ | 3.50E-03 | 0.6081 ■▼ | 3.16E-03 |
| Australian | 0.8664 ◘ ◊ | 1.47E-03 | 0.8648 ◘ ◊ | 1.56E-03 | 0.8279 ■▼ | 1.81E-03 | 0.8289 ■▼ | 1.71E-03 | 0.8289 ■▼ | 1.71E-03 | 0.8574 ◘ ◊ | 1.61E-03 | 0.8589 ◘ ◊ | 1.83E-03 |
| Blood | 0.6593 ◘ ▲ | 3.45E-03 | 0.4986 ■▼ | 2.84E-03 | 0.5363 ■▼ | 4.15E-03 | 0.5473 ■▼ | 4.12E-03 | 0.5473 ■▼ | 4.12E-03 | 0.4906 ■▼ | 2.89E-03 | 0.5119 ■▼ | 3.06E-03 |
| Bupa | 0.6628 ■▼ | 5.45E-03 | 0.6671 ◘ ▼ | 5.51E-03 | 0.6519 ■▼ | 6.45E-03 | 0.6600 ■▼ | 6.53E-03 | 0.6614 ■▼ | 6.46E-03 | 0.6290 ■▼ | 5.73E-03 | 0.6552 ■▼ | 5.92E-03 |
| Contraceptive | 0.5416 ◘ ◊ | 2.03E-03 | 0.5388 ◘ ▼ | 1.77E-03 | 0.4982 ■▼ | 1.60E-03 | 0.5119 ■▼ | 1.80E-03 | 0.4982 ■▼ | 1.45E-03 | 0.5319 ■▼ | 2.07E-03 | 0.5175 ■▼ | 1.76E-03 |
| Dermatology | 0.9748 ◘ ◊ | 7.72E-04 | 0.9733 ◘ ◊ | 7.53E-04 | 0.9282 ■▼ | 2.26E-03 | 0.9239 ■▼ | 2.86E-03 | 0.9220 ■▼ | 2.33E-03 | 0.9712 ◘ ◊ | 7.56E-04 | 0.9771 ◘ ◊ | 5.89E-04 |
| Fertility | 0.4595 ◘ ◊ | 2.69E-02 | 0.4678 ◘ ◊ | 1.37E-04 | 0.4714 ◘ ◊ | 3.06E-03 | 0.4705 ◘ ◊ | 3.11E-03 | 0.4705 ◘ ◊ | 3.11E-03 | 0.4678 ◘ ◊ | 1.37E-04 | 0.4678 ◘ ◊ | 1.37E-04 |
| Haberman | 0.6197 □▲ | 9.06E-03 | 0.5331 ■▼ | 8.74E-03 | 0.5489 ■▼ | 9.93E-03 | 0.5662 ◘ ◊ | 1.08E-02 | 0.5662 ◘ ◊ | 1.08E-02 | 0.5273 ■▼ | 7.75E-03 | 0.5112 ■▼ | 7.43E-03 |
| Heart | 0.8423 ◘ ◊ | 4.82E-03 | 0.8327 ■▼ | 4.79E-03 | 0.8028 ■▼ | 5.64E-03 | 0.7921 ■▼ | 6.76E-03 | 0.7921 ■▼ | 6.76E-03 | 0.8218 ◘ ◊ | 5.15E-03 | 0.7852 ■▼ | 5.71E-03 |
| Penbased | 0.9763 ■▼ | 1.82E-05 | 0.9693 ■▼ | 2.21E-05 | 0.9742 ■▼ | 2.18E-05 | 0.9804 ■▼ | 1.58E-05 | 0.9739 ■▼ | 2.21E-05 | 0.9586 ■▼ | 3.54E-05 | 0.9603 ■▼ | 2.95E-05 |
| Pima | 0.7289 ■▼ | 2.61E-03 | 0.7118 ■▼ | 3.69E-03 | 0.7070 ■▼ | 3.47E-03 | 0.7104 ◘ ◊ | 3.30E-03 | 0.7104 ◘ ◊ | 3.30E-03 | 0.7065 ■▼ | 3.65E-03 | 0.7119 ■▼ | 3.43E-03 |
| Plant Margin | 0.7865 ◘ ◊ | 9.21E-04 | 0.7874 ◘ ◊ | 8.98E-04 | 0.5349 ■▼ | 1.42E-03 | 0.6421 ■▼ | 1.20E-03 | 0.5336 ■▼ | 1.46E-03 | 0.7835 ◘ ◊ | 8.52E-04 | 0.7701 ■▼ | 8.84E-04 |
| Satimage | 0.6680 ■▼ | 7.91E-05 | 0.8579 ■▼ | 2.06E-04 | 0.5728 ■▼ | 1.53E-04 | 0.8631 ◘ ◊ | 1.89E-04 | 0.5728 ■▼ | 1.53E-04 | 0.8644 ◘ ◊ | 2.12E-04 | 0.8637 ■▼ | 1.96E-04 |
| Skin_NonSkin | 0.9985 ■▼ | 1.02E-07 | 0.9928 ■▼ | 4.85E-07 | 0.9992 ◘ ▲ | 4.79E-08 | 0.9992 ◘ ▲ | 4.51E-08 | 0.9992 ◘ ▲ | 4.51E-08 | 0.9528 ■▼ | 2.52E-06 | 0.9410 ■▼ | 2.97E-06 |
| Tae | 0.5403 ◘ ◊ | 1.96E-02 | 0.5281 ◘ ▼ | 1.57E-02 | 0.5536 ◘ ▲ | 1.63E-02 | 0.5153 ◘ ◊ | 1.49E-02 | 0.5599 ◘ ▲ | 1.65E-02 | 0.5166 ■▼ | 1.44E-02 | 0.5045 ◘ ◊ | 1.62E-02 |
| Texture | 0.8527 ■▼ | 1.25E-05 | 0.9856 ■▼ | 2.16E-05 | 0.9447 ■▼ | 4.00E-04 | 0.9670 ■▼ | 6.65E-05 | 0.9448 ■▼ | 4.03E-04 | 0.9862 ■▼ | 2.49E-05 | 0.9840 ■▼ | 2.75E-05 |
| Twonorm | 0.9777 ◘ ◊ | 3.27E-05 | 0.9773 ◘ ▼ | 3.33E-05 | 0.9252 ■▼ | 1.20E-04 | 0.9188 ■▼ | 1.31E-04 | 0.9188 ■▼ | 1.31E-04 | 0.9779 ◘ ▲ | 2.97E-05 | 0.9780 ■▼ | 3.03E-05 |
| Vehicle | 0.7823 ◘ ▲ | 1.37E-03 | 0.7500 ◘ ◊ | 1.88E-03 | 0.7344 ■▼ | 1.98E-03 | 0.6883 ■▼ | 2.19E-03 | 0.7228 ■▼ | 2.21E-03 | 0.7214 ■▼ | 2.52E-03 | 0.6933 ■▼ | 2.52E-03 |
| Vertebral | 0.8020 ◘ ◊ | 5.10E-03 | 0.7884 ■▼ | 5.33E-03 | 0.7570 ■▼ | 5.09E-03 | 0.7341 ■▼ | 5.26E-03 | 0.7392 ■▼ | 5.62E-03 | 0.7896 ◘ ◊ | 5.72E-03 | 0.8001 ◘ ◊ | 4.74E-03 |
| Yeast | 0.5199 ◘ ◊ | 4.07E-03 | 0.5260 ◘ ◊ | 4.34E-03 | 0.4456 ■▼ | 2.69E-03 | 0.4917 ■▼ | 4.60E-03 | 0.4339 ■▼ | 2.88E-03 | 0.5175 ◘ ◊ | 5.37E-03 | 0.4990 ■▼ | 5.68E-03 |

◘ : *The benchmark algorithm is equal to Proposed $CV_{10}$*, □: *The benchmark algorithm is better than Proposed $CV_{10}$*, ■: *The benchmark algorithm is worse than Proposed $CV_{10}$*

◊: *The benchmark algorithm is equal to Proposed $Specific_{10}$*, ▲: *The benchmark algorithm is better than Proposed $Specific_{10}$*, ▼: *The benchmark algorithm is worse than Proposed $Specific_{10}$*

## TABLE.A7. F1 SCORES AND VARIANCES OF THREE HOMOGENEOUS ENSEMBLE METHODS AND THE PROPOSED METHOD

| | Random Subspace | | AdaBoost | | Bagging | | Proposed $CV_{10}$ | | Proposed $Specific_{10}$ | |
|---|---|---|---|---|---|---|---|---|---|---|
| | Mean | Variance | Mean | Variance | Mean | Variance | Mean | Variance | Mean | Variance |
| Abalone | 0.5263 ■▼ | 5.33E-04 | 0.4272 ■▼ | 2.31E-04 | 0.5471 □▲ | 5.38E-04 | 0.5373 | 5.21E-04 | 0.5364 ◘ | 4.76E-04 |
| Artificial | 0.7071 ■▼ | 3.04E-03 | 0.7723 ■ ◊ | 2.43E-03 | 0.7801 ◘ ◊ | 2.38E-03 | 0.7857 | 2.59E-03 | 0.7776 ◘ | 2.48E-03 |
| Australian | 0.8522 ■ ◊ | 1.81E-03 | 0.8506 ■ ◊ | 1.54E-03 | 0.8629 ◘ ◊ | 1.50E-03 | 0.8649 | 1.45E-03 | 0.8583 ◘ | 1.48E-03 |

| | | | | | | | | | |
|---|---|---|---|---|---|---|---|---|---|
| Blood | 0.5747 ▪ ◊ | 3.62E-03 | 0.6440 ▫ ▲ | 3.24E-03 | 0.6326 ▫ ▲ | 2.54E-03 | 0.5822 | 3.78E-03 | 0.5728 ▪ | 3.73E-03 |
| Bupa | 0.6258 ▪ ▼ | 7.83E-03 | 0.7204 ▫ ▲ | 5.51E-03 | 0.6936 ▪ ◊ | 5.73E-03 | 0.6903 | 5.58E-03 | 0.6872 ▪ | 5.57E-03 |
| Contraceptive | 0.4505 ▪ ▼ | 1.41E-03 | 0.3740 ▪ ▼ | 8.50E-04 | 0.5228 ▪ ▼ | 1.35E-03 | 0.5342 | 1.70E-03 | 0.5332 ▪ | 2.05E-03 |
| Dermatology | 0.9718 ▪ ◊ | 6.79E-04 | 0.9559 ▪ ▼ | 1.43E-03 | 0.9524 ▪ ▼ | 1.73E-03 | 0.9754 | 7.44E-04 | 0.9737 ▪ | 7.27E-04 |
| Fertility | 0.4678 ▪ ◊ | 1.37E-04 | 0.5487 ▫ ▲ | 3.49E-02 | 0.5490 ▫ ▲ | 3.16E-02 | 0.4743 | 4.67E-03 | 0.4663 ▪ | 1.81E-04 |
| Haberman | 0.5424 ▪ ◊ | 8.89E-03 | 0.5658 ▪ ▲ | 7.71E-03 | 0.5321 ▪ ◊ | 6.53E-03 | 0.5662 | 1.08E-02 | 0.5438 ▪ | 8.95E-03 |
| Heart | 0.8235 ▪ ◊ | 5.30E-03 | 0.8095 ▪ ▼ | 6.29E-03 | 0.8116 ▪ ◊ | 5.40E-03 | 0.8405 | 4.56E-03 | 0.8281 ▪ | 5.70E-03 |
| Penbased | 0.9777 ▪ ▼ | 2.08E-05 | 0.4823 ▪ ▼ | 1.43E-04 | 0.9822 ▪ ◊ | 1.70E-05 | 0.9856 | 1.21E-05 | 0.9815 ▪ | 1.48E-05 |
| Pima | 0.6852 ▪ ▼ | 3.36E-03 | 0.7211 ▪ ◊ | 2.63E-03 | 0.7335 ▪ ▲ | 2.51E-03 | 0.7189 | 3.28E-03 | 0.7170 ▪ | 3.07E-03 |
| Plant Margin | 0.8062 ▫ ▲ | 7.12E-04 | 0.0250 ▪ ▼ | 2.00E-05 | 0.7771 ▪ ▼ | 9.65E-04 | 0.7878 | 1.00E-03 | 0.7854 ▪ | 8.28E-04 |
| Satimage | 0.8883 ▫ ▲ | 1.56E-04 | 0.7286 ▪ ▼ | 2.27E-04 | 0.8889 ▫ ▲ | 1.58E-04 | 0.8637 | 1.81E-04 | 0.8657 ▪ | 1.90E-04 |
| Skin_NonSkin | 0.9960 ▪ ▼ | 1.94E-07 | 0.9346 ▪ ▼ | 4.34E-06 | 0.9991 ▪ ◊ | 5.05E-08 | 0.9993 | 4.17E-08 | 0.9991 ▪ | 5.78E-08 |
| Tae | 0.4601 ▪ ▼ | 1.37E-02 | 0.4475 ▪ ▼ | 1.27E-02 | 0.5943 ▫ ▲ | 1.86E-02 | 0.5492 | 1.58E-02 | 0.5219 ▪ | 1.69E-02 |
| Texture | 0.9736 ▪ ▼ | 3.83E-05 | 0.5165 ▪ ▼ | 3.21E-04 | 0.9601 ▪ ▼ | 6.87E-05 | 0.9880 | 2.18E-05 | 0.9868 ▪ | 2.64E-05 |
| Twonorm | 0.9717 ▪ ▼ | 3.35E-05 | 0.9696 ▪ ▼ | 4.05E-05 | 0.9685 ▪ ▼ | 4.02E-05 | 0.9774 | 3.15E-05 | 0.9766 ▪ | 3.38E-05 |
| Vehicle | 0.7250 ▪ ▼ | 1.82E-03 | 0.4925 ▪ ▼ | 1.89E-03 | 0.7367 ▪ ▼ | 1.91E-03 | 0.7697 | 1.80E-03 | 0.7478 ▪ | 1.73E-03 |
| Vertebral | 0.6573 ▪ ▼ | 8.46E-03 | 0.5888 ▪ ▼ | 3.48E-03 | 0.7688 ▪ ▼ | 4.85E-03 | 0.7935 | 5.30E-03 | 0.7863 ▪ | 5.83E-03 |
| Yeast | 0.3447 ▪ ▼ | 3.96E-03 | 0.1384 ▪ ▼ | 2.90E-04 | 0.4655 ▪ ▼ | 2.67E-03 | 0.5254 | 4.34E-03 | 0.5221 ▪ | 4.33E-03 |

▫ : *The benchmark algorithm is equal to Proposed $CV_{10}$*, ▫: *The benchmark algorithm is better than Proposed $CV_{10}$*, ▪: *The benchmark algorithm is worse than Proposed $CV_{10}$*
◊: *The benchmark algorithm is equal to Proposed $Specific_{10}$*, ▲: *The benchmark algorithm is better than Proposed $Specific_{10}$*, ▼: *The benchmark algorithm is worse than Proposed $Specific_{10}$*